\documentclass[lettersize,journal]{IEEEtran}
\usepackage{mathrsfs}
\usepackage{amsmath,amsfonts}
\usepackage[linesnumbered,ruled,vlined]{algorithm2e}
\usepackage{array}
\usepackage{subcaption}
\usepackage[font=footnotesize]{caption}
\usepackage{multirow}
\usepackage{textcomp}
\usepackage{stfloats}
\usepackage{url}
\usepackage{verbatim}
\usepackage{graphicx}
\usepackage{cite}
\usepackage{pdflscape}
\usepackage{booktabs}
\usepackage{hyperref}
\usepackage{makecell}

\begin{document}

\title{Towards Generalization of Graph Neural Networks for AC Optimal Power Flow }

\author{\IEEEauthorblockN{Olayiwola Arowolo$^{*}$, Jochen L. Cremer$^{*}$$^{\dagger}$}\\
\IEEEauthorblockA{$^{*}$Delft University of Technology, Delft, the Netherlands \\
$^{\dagger}$Austrian Institute of Technology, Vienna, Austria \\
o.a.arowolo, j.l.cremer\{@tudelft.nl\} \\
}
}

\maketitle

\begin{abstract}
AC Optimal Power Flow (ACOPF) is computationally intensive for large-scale grids, often requiring prohibitive solution times with conventional solvers. Machine learning offers significant speedups, but existing models struggle with scalability and topology flexibility. To address these challenges, we propose a Hybrid Heterogeneous Message Passing Neural Network (HH-MPNN) that integrates a heterogeneous graph neural network (GNN) with a scalable transformer and physics-informed positional encodings. Our architecture explicitly models distinct power system components to capture local features while using global attention for long-range dependencies. Evaluated on diverse benchmarks including PGLearn and GridFM-DataKit datasets, HH-MPNN achieves \textless1\% optimality gap on default topologies across grid sizes from 14 to 2,000 buses. For N-1 contingencies, our approach demonstrates zero-shot N-1 generalization with \textless3\% optimality gap on several test cases despite training only on default topologies. We further develop an approach that ensures robust N-1 generalization to high-impact contingencies through targeted augmentation of the training data, showing that exhaustive simulation is unnecessary for topologically flexible models. Finally, size generalization experiments demonstrate that pre-training on small grids significantly improves performance on large-scale systems. Achieving computational speedups of up to 5,000 times compared to interior point solvers, these results advance practical, generalizable machine learning for real-time power system operations.
\end{abstract}

\begin{IEEEkeywords}
Graph Neural Network, Generalization, Machine Learning, Optimal Power Flow, Topology
\end{IEEEkeywords}
\vspace{-2mm}

\section{Introduction}
\IEEEPARstart{T}{he} AC optimal power flow (ACOPF) problem determines the optimal generator dispatch in a grid, subject to physical and engineering constraints \cite{cain_history_2012}. Despite being fundamental to power systems operations, including unit commitment and transmission expansion planning \cite{khaloie_review_2024}, ACOPF remains computationally challenging for large grids due to its non-convexity and non-linearity. No commercial solver guarantees globally optimal solutions for large grids within acceptable timeframes \cite{cain_history_2012}. With system operators solving ACOPF every 5-15 minutes \cite{khaloie_review_2024,cain_history_2012}, and rapid demand fluctuations from Distributed Energy Resources (DERs) requiring even faster solution times, the computational burden of conventional solvers is becoming prohibitive \cite{panciatici_operating_2012, hamann_foundation_2024}.

While modern hardware can accelerate solvers like IPOPT \cite{shin_accelerating_2024}, they remain slow for real-time use and scale poorly \cite{pan_deepopf_2023}. Operators thus rely on suboptimal relaxations \cite{yang_linearized_2018,molzahn_implementation_2013} or DC approximations \cite{chatzos_high-fidelity_2020}, which are never AC-feasible \cite{baker_solutions_2021}. Suboptimal solutions for ACOPF are particularly impactful, as even a 5\% improvement in efficiency can lead to savings of billions of dollars and significantly reduced carbon emissions \cite{cain_history_2012}.

Machine learning can accelerate ACOPF by shifting the computational burden to offline training. Existing approaches comprise: (1) end-to-end learning, which directly predicts ACOPF variables \cite{zamzam_learning_2020, park_compact_2024, huang_unsupervised_2024, owerko_unsupervised_2024} and (2) learning-to-optimize that accelerates conventional solvers through warm-starts \cite{baker_learning_2019}, active constraint predictions \cite{crozier_data-driven_2022} or ML-aided distributed optimization \cite{chatzos_spatial_2022}. See \cite{khaloie_review_2024} for a comprehensive taxonomy on ML for ACOPF. While these approaches offer varying degrees of speedups, important challenges remain regarding topology flexibility and scalability.

A key challenge is flexibility to topology changes from outages, maintenance or transmission switching. While Fully Connected Neural Networks (FNNs) are the most widely used architecture in the literature, they struggle to adapt to topology changes \cite{falconer_leveraging_2023}. Convolutional Neural Network (CNN) \cite{jia_convopf-dop_2023} and FNN  variants\cite{zhou_deepopf-ft_2023} handle limited changes in system topology, but these remain grid-specific. Graph Neural Networks (GNNs) are the most naturally suited to handling topology variations \cite{owerko_optimal_2020, gao_physics-guided_2024}, but existing approaches combine GNN feature extraction with topology-specific fully connected layers, limiting their ability to handle arbitrary topology changes. GNN approaches have also been combined with physics-informed loss to improve constraint satisfaction \cite{varbella_physics-informed_2024, lopez-garcia_optimal_2025}, or used to predict warm start points for iterative ACOPF solvers \cite{deihim_initial_2024}. However, pure GNN approaches using message passing have a 'locality bottleneck' problem. As such, they struggle to accurately predict variables like voltage angle, which depend on global grid properties, leading to significantly inferior performance compared to FNN baselines \cite{falconer_leveraging_2023,liu_topology-aware_2023}.

Existing GNN approaches also commonly use homogeneous graphs, treating all buses identically and modelling transmission lines and transformers as edges. Homogeneous graphs inadequately represent complex power networks where buses connect to varying numbers of generators, loads, and shunts. As such, homogeneous GNNs cannot be used for large, complex grids. \cite{piloto_canos_2024, ghamizi_opf-hgnn_2024} have proposed heterogeneous graph approaches for ACOPF. To overcome the 'locality bottleneck' problem, \cite{piloto_canos_2024} requires up to 60 GNN layers to propagate information through large networks, making training computationally expensive and scalability challenging. \cite{ghamizi_opf-hgnn_2024} was only evaluated on small systems (up to 30 buses) with limited complexity. Furthermore, most works require retraining/finetuning for each N-1 contingency or generating exhaustive N-1 training datasets, which becomes prohibitive for large grids. It becomes important to investigate whether a single GNN model can generalize to an unseen generator or line outage without exhaustive retraining or N-1 data generation.

Moreover, emerging research \cite{bau_principled_2025} has highlighted the limitations of generating ACOPF data through uncorrelated bus load perturbations. Though this method of load perturbation is widely used in the literature \cite{li_lumina_2026}, it leads to limited variation in total load demand and consequently, similar dispatch patterns across various ACOPF samples \cite{klamkin_pglearn_2025}. The limitation of the data generation method severely challenges the generalization assessment of several state-of-the-art ML methods for ACOPF. To gain confidence in the use of ML for ACOPF, it is important to evaluate models with challenging datasets which closely match realistic operating conditions with significant variation in total load, generator costs, network topology, etc. \cite{puech_gridfm-datakit-v1_2025}.

In this work, we address critical challenges related to topology flexibility, constraint feasibility, and scalability of ML for ACOPF. Inspired by \cite{rampasek_recipe_2023}, our novel architecture provides important insights into how power systems domain knowledge can be embedded into a graph-based model to improve performance on large-scale power systems. Our main contributions are:

\begin{enumerate}

    \item Hybrid Graph Architecture: We propose a novel architecture which integrates heterogeneous message passing, a scalable transformer, and physics-informed positional encoding. Our design overcomes the 'locality bottleneck' of standard GNNs, enabling accurate ACOPF predictions across grid sizes.
    \item Efficient N-1 Generalization: We develop an approach for robust N-1 generalization through targeted training on a small subset of high-impact contingencies, proving that exhaustive N-1 data generation is unnecessary for our topologically flexible model. We also show that our model is sufficient for accurate zero-shot prediction for a majority of N-1 contingencies.
    \item Size Generalization: We demonstrate that representations learnt on smaller, computationally "cheaper" grids can be effectively transferred to significantly larger systems (up to 2,000 buses) via fine-tuning. This offers a practical strategy to reduce the prohibitive data generation costs typically associated with large-scale grids.

\end{enumerate}
The rest of the paper is organized as follows: Section II describes our proposed model. Section III presents case studies and results. Section IV discusses the results and limitations, while Section V concludes the paper. 
\vspace{-2mm}

\footnotetext{Codes will be made public after publication}

\section{Hybrid Heterogeneous Message Passing Neural Network}
We propose a Hybrid Heterogeneous Message Passing Neural Network (HH-MPNN) for predicting ACOPF variables. MPNNs are a class of GNNs in which the model learns to make predictions on graphs through an iterative exchange of information between nodes, known as message passing. The proposed HH-MPNN consists of two components. 1) A heterogeneous MPNN which aggregates local information in the graph. 2) A scalable transformer that enables global information exchange through self-attention.
\vspace{-2mm}
\subsection{Heterogeneous Message Passing Neural Network}
Our heterogeneous MPNN follows the encode-process-decode framework \cite{battaglia_interaction_2016}. We represent the power grid with four node types (buses, generators, loads, shunts) and three edge types (transmission lines, transformers, and connector pseudo-edges linking buses to other node types). This explicit representation of power system components improves model expressivity compared to homogeneous graphs that treat all buses identically.
The architecture operates in three stages: (1) Encoding: Project node and edge features into a shared embedding space, augmented with positional encodings (see section II-B). (2) Processing: Iteratively update node and edge embeddings through message passing with residual connections to prevent over-smoothing. (3) Decoding: Map final node embeddings to predicted ACOPF variables using node type-specific decoders. The bus decoder predicts voltage angle and magnitudes for each bus, while the generator decoder predicts active and reactive power generation from each generator.

The encoding stage projects features through 2-layer multilayer perceptrons (MLPs):
\begin{subequations}
\label{eq:model_equations_1}
\begin{align} 
h^{0}_{a,i} &= E^{a}_{\theta}(x_{a,i})  \oplus PE_{i}  \label{eq:nodeencoder} \\
h^{0}_{e,i,j} &= E^{e}_{\theta}(e_{i,j}) \label{eq:edgeencoder}
\end{align}
\end{subequations}
where $a \in {[N,G,S,L]}$ denotes node type (bus, generator, shunt, load) and $e$
denotes edge type (transmission line or transformer). $x_{i,j}$ and $e_{i,j}$ are input features, $PE_{i}$ is the positional encoding and $\oplus$ denotes concatenation. Message passing updates edges using adjacent node states, then aggregates edge messages to update nodes, both with residual connections:
\begin{subequations}
\label{eq:model_equations_2}
\begin{align} 
\tilde{h}^{k}_{e,i,j} &= U^{e}_{\theta}(h_{a,i}^{k-1},h_{a,j}^{k-1} h_{e,i,j}^{k-1}) \label{eq:edgeupdater} \\
h^{k}_{e,i,j} &= h^{k-1}_{e,i,j} + \tilde{h}^{k}_{e,i,j} \label{eq:edgeresidual}\\
m^{k}_{a,i} &= \sum_{j \in N_i} (h^{k}_{e,i,j}) \label{eq:messageaggregation} \\
\tilde{h}^{k}_{a,i} &= U^{a}_{\theta}(h_{a,i}^{k-1}, m^{k}_{a,i}) \label{eq:nodeupdater} \\
h^{k}_{a,i} &= h^{k-1}_{a,i} + \tilde{h}^{k}_{a,i} \label{eq:noderesidual} 
\end{align}
\end{subequations}
All $E_{\theta}$ and $U_{\theta}$ functions are 2-layer MLPs with separate parameters for each node and edge type. For connector pseudo-edges, \ref{eq:edgeencoder} is omitted, and edge updates in \ref{eq:edgeupdater} use only adjacent node features without edge-specific features.
\vspace{-2mm}
\subsection{Transformer with Effective Resistance Positional Encoding}
To capture long-range dependencies in power grids, we augment the local MPNN with a scalable transformer using performer attention \cite{choromanski_rethinking_2021}, which approximates self-attention with linear complexity. Since graph transformers require positional encodings to distinguish node positions, we use effective resistance, a domain-informed metric that captures both local and global electrical distances as positional encoding \cite{cetinay_topological_2018}.

Effective resistance represents the potential difference between a pair of nodes when a unit of current is injected into one node and leaves at the other \cite{cetinay_topological_2018}. Under DC approximation, the power balance equation can be written as $P = Q\theta$, where $Q$ is the weighted Laplacian matrix with elements:
\begin{equation}
    Q =  
\begin{cases}
 -B_{ij}, & i \neq j \\[6pt]
 \sum_{j=1}^N B_{ij}, & i = j
\end{cases} \label{eq:graphtheory_laplacian}
\end{equation}
where $B_{ij}$ is the susceptance of the edge connecting nodes $i$ and $j$. 
From the DC power balance equation, $\theta$ can be written as $Q^{+}P$, where $Q^{+}$ is the pseudoinverse of $Q$. Effective resistance $\omega_{ij}$ can now be defined as \cite{cetinay_topological_2018}:
\begin{equation}
    \omega_{ij} = (e_i - e_j)^{T} Q^{+} (e_i - e_j) \label{eq:final_ER} \\
\end{equation} 
where $\theta_{i} - \theta_{j} = \omega_{ij}P_{ij}$ and $e_i$, $e_j$ are basic vectors with 1 at positions $i$, $j$ respectively and 0 elsewhere.

The effective resistance matrix, $\Omega \in \mathbb{R}^{N \times N}$, gives a notion of how difficult it is to move a unit of current from node $i$ to $j$, which considers all paths in the graph \cite{koc_impact_2014}. However, its dimensionality scales with grid size. To maintain a consistent dimension across different grids, we compute 5 statistical moments for each node's resistance vector:
\begin{equation}
    PE_{i} = [min, max, std, median, mean] . {\Omega}_{[i,:]} \label{eq:PSE} \\
\end{equation}
$PE_{i}$ summarizes the effective resistance from bus $i$ to other buses in the grid. This preserves distributional information about each node's electrical position while keeping the encoding dimension fixed at 5 across all grid sizes.
\begin{figure}[t!]
    \centering
    \includegraphics[width=0.48\textwidth]{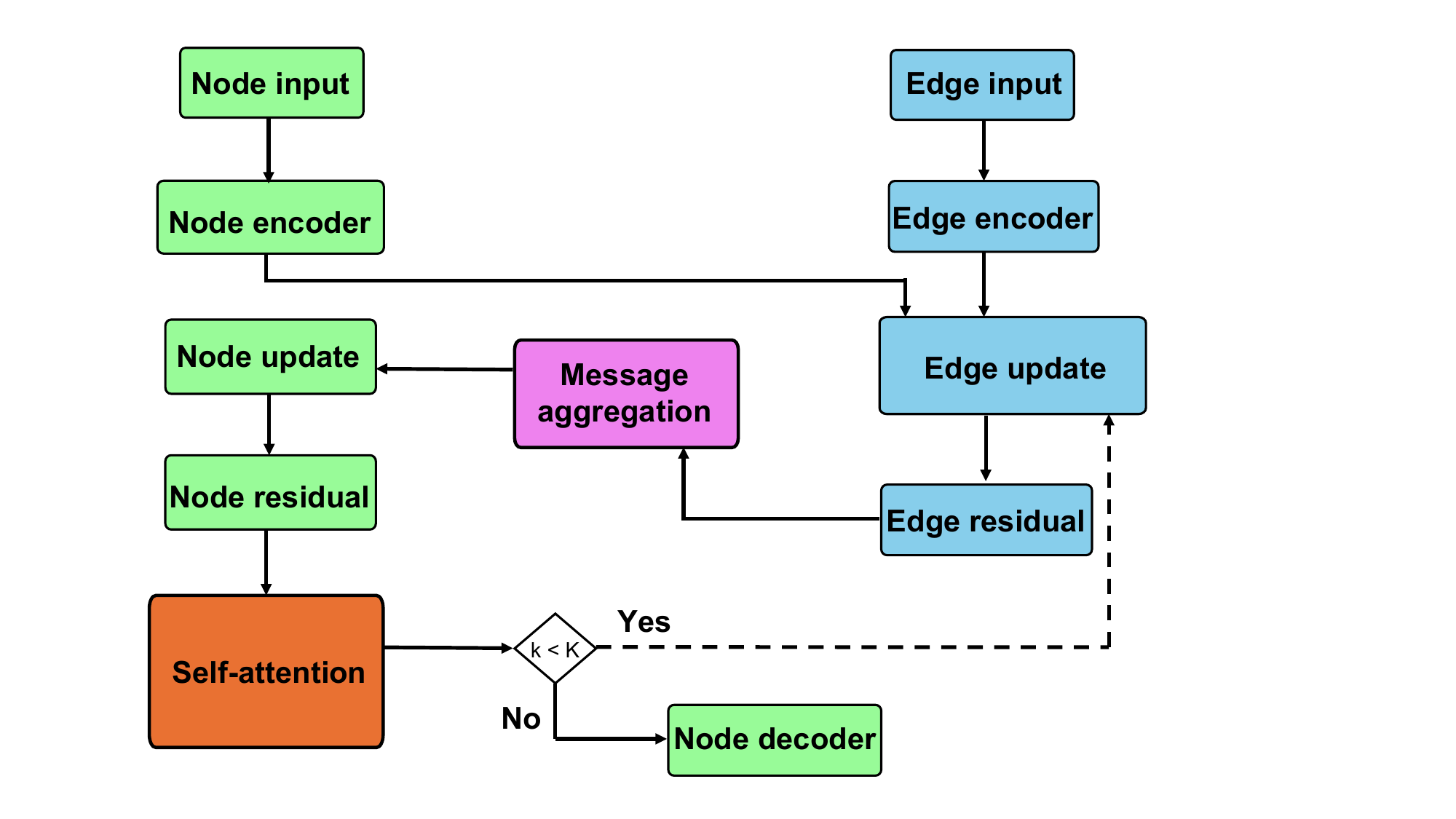}
    \caption{The proposed HH-MPNN architecture with detailed message 
    passing mechanism showing how information flows between different 
    node and edge types. Skip connections and bidirectional updates enhance model expressivity. K=5 message passing layers are used.}
    \label{fig:model_architecture}
\end{figure}
Algorithm \ref{alg:HH-MPNN} summarises the complete HH-MPNN forward pass. After encoding node features with positional information in line \ref{alg1-l1}, we alternate between local message passing (line \ref{alg1-l3}) and global attention (line \ref{alg1-l5}) for $K$ layers. The attention mechanism operates on concatenated node embeddings from all types (line \ref{alg1-l4}), enabling information exchange across the entire graph. We combine local MPNN updates with global transformer updates via element-wise addition (line \ref{alg1-l7}), with a 2-layer MLP projection $T_{\theta}$ to integrate the representations. The transformer output also employs a residual connection (line \ref{alg1-l6}). Type-specific 2-layer MLP decoders (buses and generators) produce final predictions (line \ref{alg1-l9}). We use K=5 message passing layers in all experiments.
\begin{algorithm}[t]
  \caption{Algorithm for HH-MPNN}
  \label{alg:HH-MPNN}
  $X^{0} = E^{a}_{\theta}(X) \oplus PE$ \label{alg1-l1} \\
  \For{$k = 0, 1, \ldots, K-1$}{ \label{alg1-l2}
  $(X^{k+1}_M, E^{k+1}) = \text{MPNN}(X^k, E^k)$ \label{alg1-l3} \\ 
  \textit{Aggregate node embeddings for all node types} \label{alg1-l4} \\
  $\hat{X}^{k+1}_T = \text{AT}(X^k)$ \label{alg1-l5} \\
  $X^{k+1}_T = X^k_T + \hat{X}^{k+1}_T$ \label{alg1-l6} \\
  $X^{k+1} = T_{\theta}(X^{k+1}_T + X^{k+1}_M)$ \label{alg1-l7} \\
  \textit{Disaggregate node embeddings by node types} \label{alg1-l8} \\
  }
  $Y = D^{a}_{\theta}(X^K)$ \label{alg1-l9} \\
\end{algorithm}
\vspace{-2mm}
\section{Case study}
\vspace{-1mm}
\subsection{Dataset}
We use 3 open-source ACOPF datasets for validating our approach. The OPFData \cite{lovett_opfdata_nodate} and PGLearn \cite{klamkin_pglearn_2025} datasets are publicly available, and we generate the Datakit dataset with the GridFM-datakit tool \cite{puech_gridfm-datakit-v1_2025}. Each dataset has two variants: 'full topology' dataset with the default grid topology and 'N-1 topology' dataset, which has a randomly disconnected component (branch or generator). Where available, we use 6 grids from each dataset for our experiments: 14, 30, 57, 118, 500 and 2000-bus grids. 
In addition to N-1 topology variations, GridFM-datakit allows variation in generator costs and network parameters, potentially increasing the diversity of dispatch patterns. We provide details on the settings we use for GridFM-datakit data generation in the appendix \ref{appendix-C}.
\vspace{-2mm}
\subsection{Baselines}
We compare the performance of our proposed approach with the following baseline machine learning models:

\subsubsection{Fully connected neural network (FNN)}
The network comprises 5 linear layers with a hidden dimension of 256 and ReLU activations. Input and output vectors are flattened, making the architecture tied to a specific grid topology.

\subsubsection{Convolutional neural network (CNN)}
We use 5 1D convolutional layers with ReLU activation and max pooling, followed by adaptive average pooling and a linear output layer with sigmoid activation. Like FNN, this architecture is topology-specific.

\subsubsection{Graph convolutional neural network (GCN)} 
\begin{figure}[ht]
    \centering
    \begin{subfigure}[b]{0.45\textwidth}
        \centering
        \includegraphics[width=\textwidth, trim=0 20mm 0 20mm, clip]{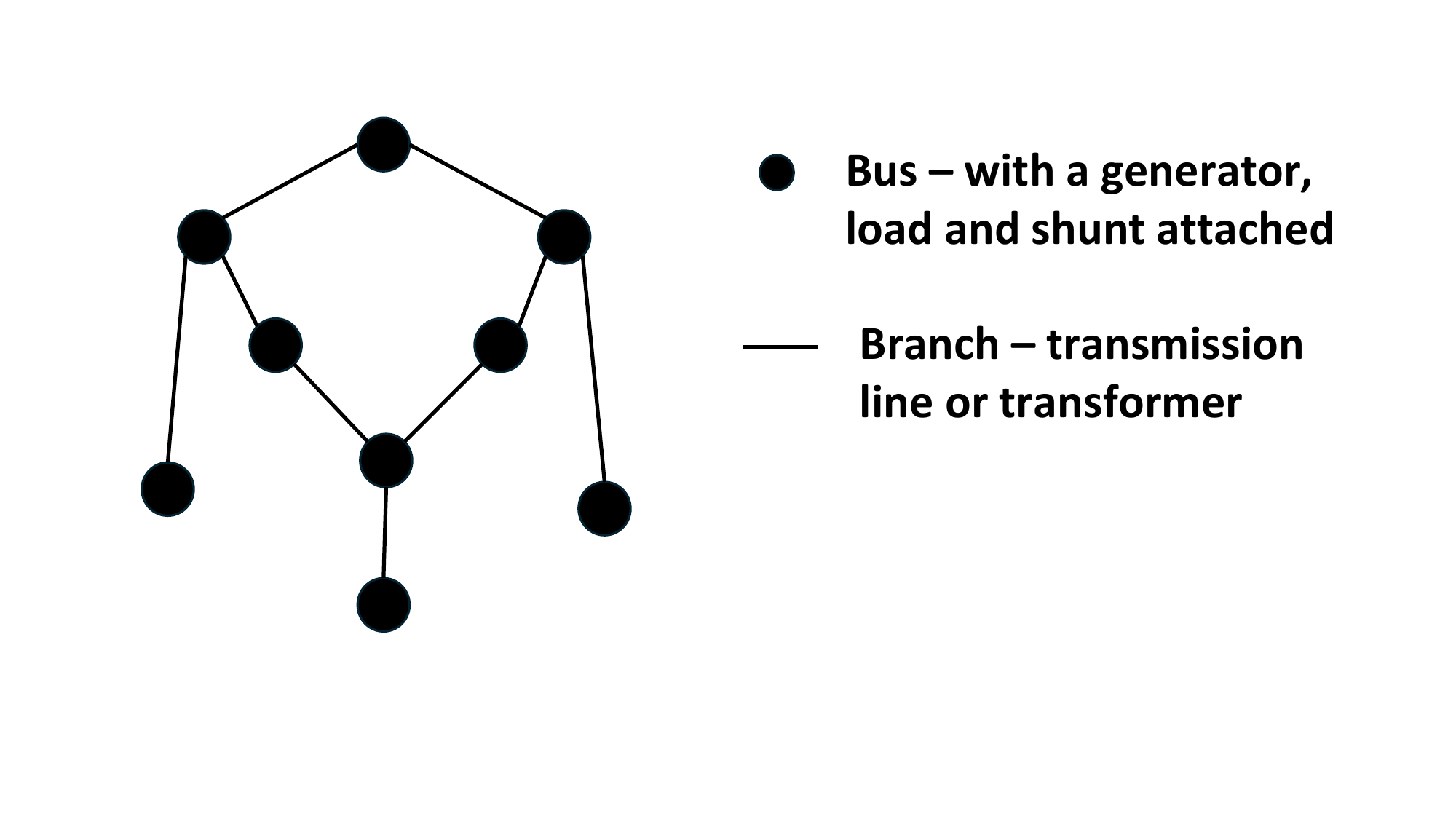}
        \caption{Homogeneous representation of the power grid models transmission lines and transformers as the same. It is difficult to represent different numbers of components attached to each bus.}
        \label{fig:homogenous_graph}
    \end{subfigure}
    \hfill
    \begin{subfigure}[b]{0.45\textwidth}
        \centering
        \includegraphics[width=\textwidth, trim=0 10mm 0 10mm, clip]{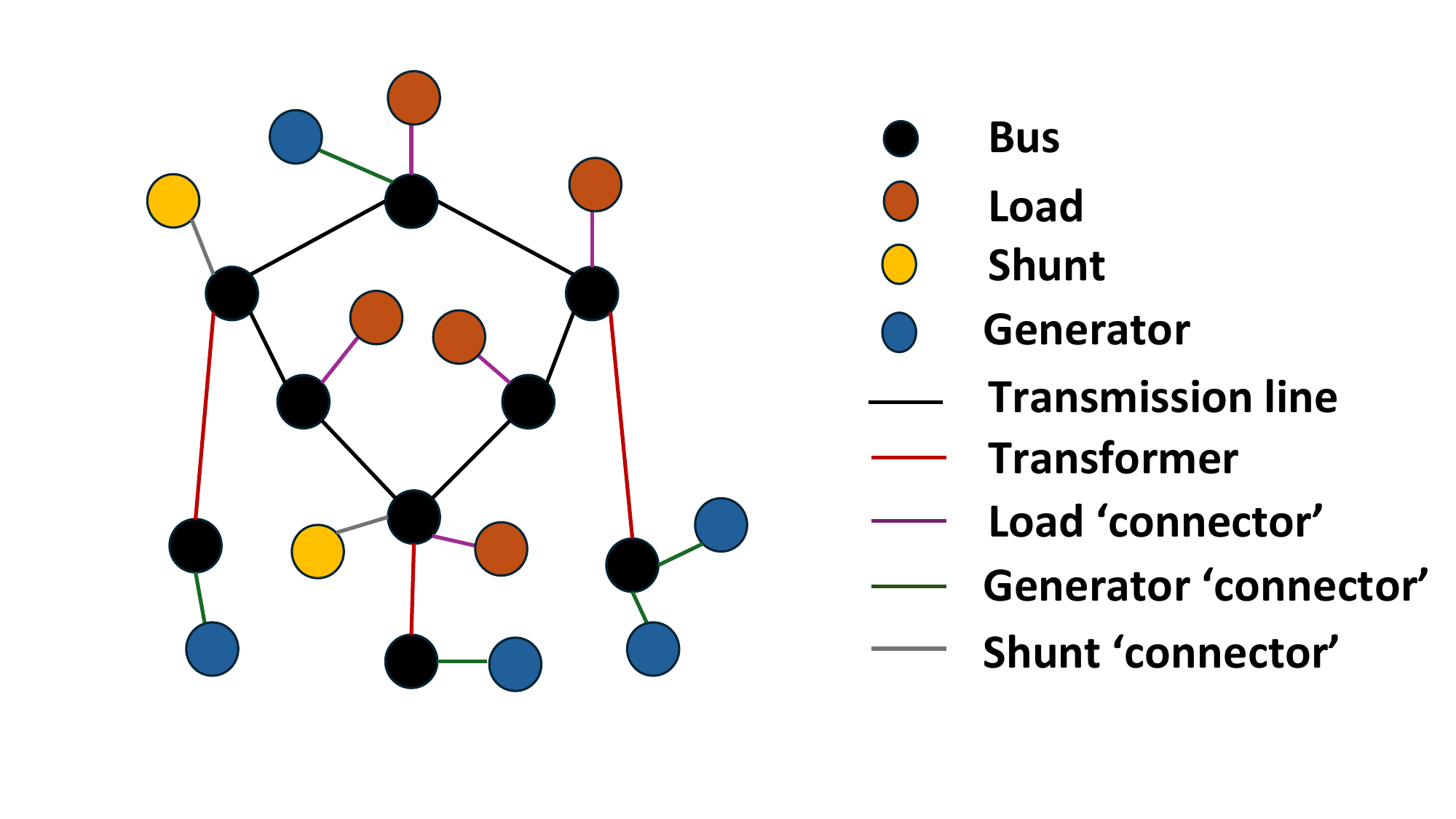}
        \caption{Proposed heterogeneous graph with different node and edge types enables a more accurate representation of the power grid.}
        \label{fig:heterogeneous_graph}
    \end{subfigure}
    \caption{Heterogeneous versus Homogeneous graph representation of the power grid}
    \label{fig:homogeneousvsheterogeneous}
\end{figure}

We represent buses as nodes and branches as edges, assuming uniform components at each bus (generator, load, shunt) and using generator masks to indicate presence/absence. This homogeneous representation is less flexible than our heterogeneous approach, as shown in Fig.\ref{fig:homogeneousvsheterogeneous}. The model comprises 5 graph convolution layers with a hidden dimension of 256. We apply GCN only to grids for which homogeneous representation is tractable (14, 30, 57, and 118-bus systems). In the case of 500 and 2000-bus systems, multiple generators with different costs may be connected to a bus, making a homogeneous representation difficult. 
\vspace{-4mm}
\subsection{Experiment settings}
We predict the voltage angles $\theta$, voltage magnitude $V$, active generator power $PG$ and reactive generator power $QG$. Following \cite{pan_deepopf_2023}, we apply a sigmoid function to bound outputs and satisfy inequality constraints (see Appendix \ref{appendix-A}, \ref{eq:pgbound} - \ref{eq:vbound}). Branch flows are computed using power flow equations (\ref{eq:forwardflow}-\ref{eq:reverseflow}) after the model prediction. Angle difference \ref{eq:thetabound}, branch thermal limit \ref{eq:flowbound}, and power balance \ref{eq:powerbalance} constraints may be violated. We train the models with supervised mean square error (MSE) loss between predicted and ground truth variables. All models use Adam optimizer with learning rate 1 x $10^{-5}$ and weight decay 5 x $10^{-8}$, trained for 100 epochs. We use  270,000 samples for training ($90\%$), 15,000 for validation ($5\%$) and 15,000 for testing ($5\%$) for the OPFData and GridFM datasets. For the PGLearn datasets, we also train on 270,000 samples, but we evaluate on the entire test dataset provided by the dataset. Batch size is 256 for small grids (14-118 buses) and 64 for large grids (500-2,000 buses). All models have 5 layers with a hidden dimension of 256. All experiments were performed on a Linux machine with 32 CPU cores and 256GB of memory and a NVIDIA A30 GPU with 24GB of memory.
\vspace{-4mm}
\subsection{Metrics}
We evaluate models using three metrics: (1) \textbf{Mean Squared Error (MSE)} between predictions and ground truth for each ACOPF variable. (2) \textbf{Optimality gap}: percentage difference between model and ground truth objective costs. (3) \textbf{Constraint satisfaction}: average and maximum violations of constraints not strictly enforced (\ref{eq:thetabound},\ref{eq:flowbound},\ref{eq:powerbalance}). Generator power and voltage magnitude constraints (\ref{eq:pgbound} - \ref{eq:vbound}) are satisfied by design through sigmoid activation. We report constraint violations over the whole test dataset.
\begin{table*}[!t]
\centering
\renewcommand{\arraystretch}{1.2}
\begin{tabular}{c l | c c c c | c c c c | c c c c}
\hline\hline
\multirow{2}{*}{\textbf{System}} & \multirow{2}{*}{\textbf{Model}} 
& \multicolumn{4}{c|}{\textbf{OPFData ($\times 10^{-4}$)}} 
& \multicolumn{4}{c|}{\textbf{PGLearn ($\times 10^{-4}$)}} 
& \multicolumn{4}{c}{\textbf{DataKit ($\times 10^{-4}$)}} \\
\cline{3-14}
& & $\theta$ & $V$ & $P_G$ & $Q_G$ 
  & $\theta$ & $V$ & $P_G$ & $Q_G$ 
  & $\theta$ & $V$ & $P_G$ & $Q_G$ \\
\hline
\multirow{4}{*}{14-IEEE} 
  & FNN      & 0.40 & 0.02 & 3.00 & 3.00 & 0.42 & 0.18 & 11.0 & 3.0 & 0.35 & 0.09 & 2.00 & 5.00\\
  & CNN      & 0.03 & 0.02 & \textbf{0.08} & 2.00 & 0.10 & 0.04 & 0.50 & 0.94 & 0.09 & 0.07 & 1.00 & 3.00 \\
  & GCN      & 3.00 & 0.43 & 17.00 & 3.00 & 6.00 & 2.00 & 72.00 & 13.00 & 3.00  & 1.00 & 50.00 & 11.00 \\
  & HH-MPNN  & \textbf{0.01} & \textbf{0.00} & 0.21 & \textbf{0.06}  & \textbf{0.01} & \textbf{0.01} & \textbf{0.04} & \textbf{0.04} & \textbf{0.06} & \textbf{0.01} & \textbf{0.12} & \textbf{0.05} \\
\hline
\multirow{4}{*}{30-IEEE} 
  & FNN      & 0.32 & 0.06 & 13.00 & 3.00 & 0.27 & 0.06 & 3.00 & 4.00 & 0.21 & 0.08 & 2.00 & 6.00 \\
  & CNN      & 0.07 & 0.05 & \textbf{0.32}  & 2.00 & 0.09 & 0.03 & 0.41 & 3.0 & 0.08 & 0.05 & 1.00 & 3.00\\
  & GCN      & 2.00 & 0.60 & 3.00  & 0.98 & 5.00 & 1.00 & 28.00 & 5.00 & 2.00 & 0.75 & 15.00 & 3.00\\
  & HH-MPNN  & \textbf{0.05} & \textbf{0.01} & 0.37  & \textbf{0.24} & \textbf{0.01} & \textbf{0.01} & \textbf{0.02} & \textbf{0.13} & \textbf{0.06} & \textbf{0.01} & \textbf{0.01} & \textbf{0.19} \\
\hline
\multirow{4}{*}{57-IEEE} 
  & FNN      & 0.63 & 0.06 & 142.00 & 44.00 & 0.26 & 0.06 & 43.0 & 22.0 & 0.43 & 0.11 & 18.0 & 26.0 \\
  & CNN      & 0.64 & 0.06 & 139.00 & 44.00 & 0.20 & 0.04 & 9.00 & 16.00 & 0.34 & 0.09 & 29.00 & 23.00 \\
  & GCN      & 3.00 & 1.00 & \textbf{60.00}  & \textbf{6.00} & 3.00 & 1.00 & 47.00 & 11.00 & 31.00 & 3.00 & 408.00 & 34.00 \\
  & HH-MPNN  & \textbf{0.37} & \textbf{0.02} & 106.00 & 20.00 & \textbf{0.03} & \textbf{0.01} & \textbf{4.00} & \textbf{0.84} & \textbf{0.15} & \textbf{0.03} & \textbf{6.00} & \textbf{3.00} \\
  \hline
\multirow{4}{*}{118-IEEE} 
  & FNN      & 2.00 & 0.10 & 55.00 & 75.00 & 3.0 & 0.07 & 37.00 & 35.00 & 7.00 & \textbf{0.44} & 273.00 & 156.00 \\
  & CNN      & \textbf{0.83} & \textbf{0.04} & \textbf{12.00} & \textbf{22.00} & \textbf{0.58} & \textbf{0.05} & \textbf{7.00} & 20.00 & 21.00 & 0.58 & 277.00 & 238.00 \\
  & GCN      & 9.00 & 0.77 & 318.00 & 68.00 & 26.00 & 1.00 & 479.00 & 246.00 & 186.00 & 4.00 & 2167.00 & 585.00 \\
  & HH-MPNN  & 3.00 & 0.07 & 42.00 & 33.00  & 0.85 & 0.07 & \textbf{7.00} & \textbf{11.00} & \textbf{6.00} & 0.57 & \textbf{79.00} & \textbf{96.00} \\
\hline
\multirow{4}{*}{500-GOC} 
  & FNN      & 3.00 & 0.05 & 13.00 & 11.00 & N/A & N/A & N/A & N/A & 4.00 & 0.11 & 30.00 & 24.00 \\
  & CNN      & 3.00 & 0.05 & 13.00 & 11.00 & N/A & N/A & N/A & N/A & 9.00 & 0.43 & 440.00 & 88.00 \\
  & GCN      & N/A  & N/A  & N/A   & N/A   & N/A & N/A & N/A & N/A & N/A & N/A & N/A & N/A \\
  & HH-MPNN  & \textbf{1.00} & \textbf{0.03} & \textbf{6.00}  & \textbf{7.00} & N/A & N/A & N/A & N/A & \textbf{1.57} & \textbf{0.22} & \textbf{7.63} & \textbf{17.64} \\
\hline
\multirow{4}{*}{2000-GOC} 
  & FNN      & 0.88 & 0.09 & 10.00 & 17.00 & N/A & N/A & N/A & N/A & 3.00 & 0.27 & 62.00 & 42.00 \\
  & CNN      & 6.00 & 0.09 & 10.00 & 17.00  & N/A & N/A & N/A & N/A & 12.00 & 0.46 & 141.00 & 69.00 \\
  & GCN      & N/A  & N/A  & N/A   & N/A   & N/A & N/A & N/A & N/A & N/A & N/A & N/A & N/A \\
  & HH-MPNN  & \textbf{0.34} & \textbf{0.06} & \textbf{5.00} & \textbf{9.00} & N/A & N/A & N/A & N/A & \textbf{2.03} & \textbf{0.34} & \textbf{16.05} & \textbf{18.76} \\
\hline\hline
\end{tabular}
\caption{Mean Squared Error (MSE) of the models on the full topology dataset. Errors are in pu except for $\theta$ in radians. The best (lowest) error for each dataset is highlighted.}
\label{tab:predicted_mse_results}
\end{table*}

\begin{table*}[t]
\centering
\renewcommand{\arraystretch}{1.2}
\begin{tabular}{ll|cccc|cccc|c}
\hline\hline
\multirow{2}{*}{System} & \multirow{2}{*}{Model}
  & \multicolumn{4}{c|}{Average constraint violation ($\times 10^{-4}$ p.u.)}
  & \multicolumn{4}{c|}{Max constraint violation (p.u.)}
  & \multirow{2}{*}{Optimality gap (\%)} \\
 & & $S_{ij}(+)$ & $S_{ji}(-)$ & $P_b$ & $Q_b$
   & $S_{ij}(+)$ & $S_{ji}(-)$ & $P_b$ & $Q_b$ & \\
\hline
\multirow{4}{*}{14-IEEE}
  & FNN     & 0.00 & 0.00 & 6.00 & 8.00 & 0.00 & 0.00 & 0.41 & \textbf{0.14} & 0.68 \\
  & CNN     & 0.00 & 0.00 & \textbf{0.16} & \textbf{3.00}  & 0.00 & 0.00 & \textbf{0.37} & \textbf{0.14} & \textbf{0.03} \\
  & GCN     & 0.00 & 0.00 & 6.00 & 17.00 & 0.00 & 0.00 & 1.05 & 0.46 & 2.84 \\
  & HH-MPNN & 0.00 & 0.00 & 0.29 & 1.15  & 0.00 & 0.00 & 0.68 & 0.16 & 0.05 \\
\hline
\multirow{4}{*}{30-IEEE}
  & FNN     & 1.00 & \textbf{0.00} & 5.00 & \textbf{0.30}  & 0.10 & 0.10 & 0.49 & 0.22 & 1.81 \\
  & CNN     & \textbf{0.25}  & 0.07 & \textbf{0.04} & 1.00  & \textbf{0.08} & \textbf{0.06} & \textbf{0.38} & \textbf{0.18} & \textbf{0.01} \\
  & GCN     & 188.00  & 186.00  & 18.00 & 27.00   & 0.79 & 0.78 & 1.81 & 1.17 & 7.28 \\
  & HH-MPNN & 3.44  & 0.30  & 0.36 & 2.00 & 0.09 & \textbf{0.06} & 0.63 & 0.27 & 0.14 \\
\hline
\multirow{4}{*}{57-IEEE}
  & FNN     & 0.00 & 0.00 & \textbf{0.05} & 2.00  & 0.00 & 0.00 & 1.00 & 0.67 & \textbf{0.08} \\
  & CNN     & 0.00  & 0.00 & 3.00 & \textbf{0.43}  & 0.00 & 0.00 & \textbf{0.85} & \textbf{0.50} & 0.22 \\
  & GCN     & 0.00  & 0.00  & 2.00 & 15.00 & 0.00 & 0.00 & 3.58 & 1.46 & 0.99 \\
  & HH-MPNN & 0.00  & 0.00  & 2.85 & 0.70 & 0.00 & 0.00 & 2.57 & 0.68 & 0.15 \\
\hline
\multirow{4}{*}{118-IEEE}
  & FNN     & \textbf{2.00} & \textbf{2.00} & \textbf{0.86} & \textbf{4.00}  & \textbf{1.97} & \textbf{1.97} & \textbf{6.37} & \textbf{1.21} & \textbf{0.04} \\
  & CNN     & 5.00  & 5.00 & 2.00 & \textbf{4.00}  & 3.10 & 3.10 & 10.40 & 1.58 & \textbf{0.04} \\
  & GCN     & 925.00  & 924.00  & 255.00 &  514.00  & 26.72 & 26.87 & 30.52 & 9.69 & 21.44 \\
  & HH-MPNN & 9.00  & 10.13  & 0.90 & 22.70 & 3.23 & 3.26 & 7.34 & 3.10 & 0.32 \\
\hline\hline
\end{tabular}
\caption{Constraint violations and optimality gap for PGLearn Dataset. The best (lowest) violations and optimality gaps are highlighted.
}
\label{tab:con_vio_and_optimality_pglearn}
\end{table*}
\vspace{-2mm}
\subsection{Prediction on 'full topology' dataset}
Tables \ref{tab:predicted_mse_results}-\ref{tab:con_vio_and_optimality_datakit} summarize the results on the full topology datasets. \textbf{Prediction accuracy (Table \ref{tab:predicted_mse_results})}: We make three observations about the results. First, model errors generally increase with increasing grid size. The pattern is consistent across the three datasets, suggesting that larger grids with more generators have more complex dispatch patterns, which make learning more challenging. However, model performance does not decrease monotonically; for example, models achieve higher accuracy on the 2000-bus system than on the 118-bus system for the OPFData and Datakit datasets. 
Second, the GCN model consistently has the worst performance, with orders of magnitude larger errors on the 118-bus system compared to other models. This highlights the poor scaling behaviour of the baseline graph model and its difficulty with predicting generator dispatches, which have little to no locality dependence. HH-MPNN provides significant improvements over the GCN model, with the hybrid architecture achieving similar or better accuracy compared to the global architectures (FNN and CNN) even on the largest dataset. 
Third, the same model trained on different datasets can achieve accuracy of different orders of magnitude. Across the 3 datasets, similar error values can be observed on the smaller grids (14, 30 and 57-bus), but on the larger datasets (500 and 2000-bus), errors of different orders of magnitude can be observed between the OPFData and the Datakit datasets. This confirms the concern in the literature that the data sampling method (particularly correlated versus uncorrelated loads) strongly influences machine learning benchmark outcomes.  The Datakit dataset is the most diverse of the three and constitutes the most challenging learning problem for the machine learning models. Interestingly, our proposed HH-MPNN model achieves the lowest error for all variables and for all the tested grids on the Datakit dataset, highlighting the superior performance of our approach. 

\textbf{Constraint violations and optimality gap (Table \ref{tab:con_vio_and_optimality_pglearn}-\ref{tab:con_vio_and_optimality_datakit})}: We also present the optimality gap and constraint violations of the models since the predicted solutions are often infeasible. All models satisfy angle difference constraints ($\theta_{ij}$) completely, so we do not include these in the constraint violations. Branch flow limits and power balance are the only remaining constraints that can be violated. First, there are no branch flow violations for the 14 and 57-bus systems across models and datasets, which suggests that lines are not congested in these systems. Once again, we see the GCN model incurs orders of magnitude higher line violations than the other models on the 30-bus and 118-bus systems, a pattern which holds across datasets. However, the HH-MPNN model frequently obtains lower average and maximum constraint violations than the FNN and CNN models. The performance of the HH-MPNN is also remarkably consistent, with optimality gaps $<1\%$ on all systems and datasets. The superior performance of the model is more apparent on the most complex Datakit dataset, where generator costs are not fixed. Here, the HH-MPNN model achieves superior or equal optimality gaps while maintaining comparable violations to the global models. Power balance constraint violations in the order of $10^{-4}$ to $10^{-2}$ are the most significant violations incurred by the model. But this is acceptable for approximate solvers and comparable to state-of-the-art results \cite{pan_deepopf_2023, piloto_canos_2024}. 

\begin{table*}[t]
\centering
\renewcommand{\arraystretch}{1.2}
\begin{tabular}{ll|cccc|cccc|c}
\hline\hline
\multirow{2}{*}{System} & \multirow{2}{*}{Model}
  & \multicolumn{4}{c|}{Average constraint violation ($\times 10^{-4}$ p.u.)}
  & \multicolumn{4}{c|}{Max constraint violation (p.u.)}
  & \multirow{2}{*}{Optimality gap (\%)} \\
 & & $S_{ij}(+)$ & $S_{ji}(-)$ & $P_b$ & $Q_b$
   & $S_{ij}(+)$ & $S_{ji}(-)$ & $P_b$ & $Q_b$ & \\
\hline
\multirow{4}{*}{14-IEEE}
  & FNN     & 0.00 & 0.00 & 5.60 & 0.56 & 0.00 & 0.00 & 0.12 & 0.08 & 0.28 \\
  & CNN     & 0.00 & 0.00 & \textbf{0.21} & 1.22 & 0.00 & 0.00 & 0.11 & 0.06 & \textbf{0.01} \\
  & GCN     & 0.00 & 0.00 & 8.08 & 40.13 & 0.00 & 0.00 & 0.90 & 0.46 & 0.67 \\
  & HH-MPNN & 0.00 & 0.00 & 0.53 & \textbf{0.22}  & 0.00 & 0.00 & \textbf{0.10} & \textbf{0.05} & 0.07 \\
\hline
\multirow{4}{*}{30-IEEE}
  & FNN     & \textbf{0.00} & \textbf{0.00} & 0.56 & 0.65  & \textbf{0.00} & \textbf{0.00} & \textbf{0.21} & \textbf{0.08} & 0.36 \\
  & CNN     & 4.42 & 0.05 & \textbf{0.14} & 1.14  & \textbf{0.00} & \textbf{0.00} & 0.27 & 0.10 & \textbf{0.06} \\
  & GCN     & 226.50 & 217.70 & 8.53 & 36.29 & 0.92 & 0.88 & 1.89 & 0.84 & 0.61 \\
  & HH-MPNN & 0.25 & 0.00 & 0.19 & \textbf{0.62} & 0.04 & 0.00 & 0.32 & 0.12 & 0.13 \\
\hline
\multirow{4}{*}{57-IEEE}
  & FNN     & 0.00 & 0.00 & \textbf{1.68} & \textbf{0.46} & 0.00 & 0.00 & \textbf{0.36} & \textbf{0.18} & 0.13 \\
  & CNN     & 0.00 & 0.00 & 1.83 & 3.26  & 0.00 & 0.00 & 0.44 & 0.25 & \textbf{0.08} \\
  & GCN     & 0.00 & 0.00 & 4.08 & 4.73 & 0.00 & 0.00 & 3.03 & 1.41 & 0.36 \\
  & HH-MPNN & 0.00 & 0.00 & 5.51 & 4.50  & 0.00 & 0.00 & 0.91 & 0.41 & 0.25 \\
\hline
\multirow{4}{*}{118-IEEE}
  & FNN     & \textbf{0.28} & \textbf{0.06} & 6.86 & \textbf{1.53} & \textbf{0.47} & \textbf{0.47} & \textbf{2.61} & \textbf{0.67} & 0.28 \\
  & CNN     & 1.57 & 1.18 & \textbf{0.15} & 5.60 & 1.17 & 1.16 & 3.71 & 1.28 & \textbf{0.01} \\
  & GCN     & 477.10 & 478.70 & 225.10 & 310.00 & 4.98 & 5.01 & 8.33 & 5.62 & 9.94 \\
  & HH-MPNN & 11.12 & 11.35 & 3.04 & 13.75  & 2.44 & 2.45 & 4.39 & 1.89 & 0.05 \\
\hline
\multirow{4}{*}{500-GOC}
  & FNN     & 3.65 & 3.45 & 1.47 & \textbf{0.88} & 0.31 & 0.30 & 5.12 & 1.20 & 0.05 \\
  & CNN     & \textbf{0.00} & \textbf{0.00} & \textbf{1.18} & 2.34  & \textbf{0.00} & \textbf{0.00} & \textbf{3.21} & \textbf{0.64} & \textbf{0.04} \\
  & GCN     & N/A  & N/A  & N/A & N/A  & N/A & N/A & N/A & N/A & N/A \\
  & HH-MPNN & 48.32 & 48.42 & 3.38 & 37.55 & 19.55 & 19.50 & 66.16 & 17.59 & 0.17 \\
\hline
\multirow{4}{*}{2000-GOC}
  & FNN     & \textbf{0.00} & \textbf{0.00} & \textbf{0.73} & \textbf{1.36} & \textbf{0.00} & \textbf{0.00} & 3.27 & 0.71 & 0.02 \\
  & CNN     & \textbf{0.00} & \textbf{0.00} & 1.26 & 8.48 & \textbf{0.00} & \textbf{0.00} & \textbf{2.49} & \textbf{0.60} & \textbf{0.01} \\
  & GCN     & N/A  & N/A  & N/A & N/A   & N/A & N/A & N/A & N/A & N/A \\
  & HH-MPNN & 0.85 & 0.86 & 2.19 & 9.59 & 3.84 & 3.84 & 24.80 & 8.64 & \textbf{0.01} \\
\hline\hline
\end{tabular}
\caption{Constraint violations and optimality gap for OPFData Dataset. The best (lowest) violations and optimality gaps are highlighted.
}
\label{tab:con_vio_and_optimality_opfdata}
\end{table*}

\begin{table*}[ht!]
\centering
\renewcommand{\arraystretch}{1.2}
\begin{tabular}{ll|cccc|cccc|c}
\hline\hline
\multirow{2}{*}{System} & \multirow{2}{*}{Model}
  & \multicolumn{4}{c|}{Average constraint violation ($\times 10^{-4}$ p.u.)}
  & \multicolumn{4}{c|}{Max constraint violation (p.u.)}
  & \multirow{2}{*}{Optimality gap (\%)} \\
 & & $S_{ij}(+)$ & $S_{ji}(-)$ & $P_b$ & $Q_b$
   & $S_{ij}(+)$ & $S_{ji}(-)$ & $P_b$ & $Q_b$ & \\
\hline
\multirow{4}{*}{14-IEEE}
  & FNN     & 0.00 & 0.00 & 2.00 & \textbf{0.74}  & 0.00 & 0.00 & 0.50 & 0.50 & 0.23 \\
  & CNN     & 0.00 & 0.00 & \textbf{0.20} & 3.00  & 0.00 & 0.00 & \textbf{0.38} & \textbf{0.38} & \textbf{0.04} \\
  & GCN     & 0.00 & 0.00 & 8.00 & 12.00 & 0.00 & 0.00 & 1.26 & 0.76 & 3.77 \\
  & HH-MPNN & 0.00 & 0.00 & 0.72 & 4.00  & 0.00 & 0.00 & 0.42 & 0.30 & 0.15 \\
\hline
\multirow{4}{*}{30-IEEE}
  & FNN     & \textbf{0.20} & 0.35 & 3.00 & \textbf{2.00}  & 0.52 & 0.52 & 0.79 & 0.48 & 1.11 \\
  & CNN     & 1.00  & \textbf{0.13} & 3.0 & \textbf{2.00}  & \textbf{0.18} & \textbf{0.18} & \textbf{0.51} & 0.35 & 0.53 \\
  & GCN     & 199.00  & 190.00  & 3.00 & 26.00   & 1.11 & 1.07 & 2.13 & 1.05 & 5.73 \\
  & HH-MPNN & 3.00  & 1.00  & \textbf{0.55} & 3.00 & 0.28 & 0.24 & 0.58 & \textbf{0.29} & \textbf{0.13} \\
\hline
\multirow{4}{*}{57-IEEE}
  & FNN     & 0.00 & 0.00 & 2.00 & \textbf{1.00}  & 0.10 & 0.00 & 2.09 & 1.72 & 0.23 \\
  & CNN     & 0.00  & 0.00 & \textbf{0.01} & 5.00  & 0.05 & 0.00 & 1.74 & 0.87 & 0.11 \\
  & GCN     & 0.00  & 0.00  & 44.00 & 2.00 & 0.00 & \textbf{0.00} & 5.33 & 1.52 & 7.05 \\
  & HH-MPNN & 0.02  & 0.00  & 2.00 & 5.00 & 0.36 & 0.04 & \textbf{1.28} & \textbf{0.66} & \textbf{0.10} \\
\hline
\multirow{4}{*}{118-IEEE}
  & FNN     & 208.00 & 209.00 & 33.00 & 79.00  & 26.29 & 26.27 & 40.84 & 8.43 & 0.30 \\
  & CNN     & \textbf{27.00}  & 29.00 & \textbf{29.00} & \textbf{18.00}  & \textbf{3.42} & \textbf{3.43} & \textbf{12.36} & \textbf{3.81} & 1.16 \\
  & GCN     & 3071.00  & 3079.00  & 260.00 & 1266.00   & 19.69 & 19.72 & 26.77 & 10.79 & 8.55 \\
  & HH-MPNN & 38.24  & 39.23  & \textbf{3.91} & 58.20 & 4.89 & 4.90 & 16.90 & 4.53 & \textbf{0.10} \\
\hline
\multirow{4}{*}{500-GOC}
  & FNN     & \textbf{73.00} & \textbf{73.00} & \textbf{0.87} & \textbf{64.00}  & 42.32 & 42.33 & 188.91 & 52.68 & 0.60 \\
  & CNN     & 123.00  & 123.00 & 26.00 & 114.00  & \textbf{10.70} & \textbf{10.70} & 62.16 & \textbf{17.34} & 1.15 \\
  & GCN     & N/A  & N/A  & N/A & N/A  & N/A & N/A & N/A & N/A & N/A \\
  & HH-MPNN & 104.18  & 104.19  & 24.08 & 81.10 & 28.54 & 28.70 & \textbf{53.11} & 22.52 & \textbf{0.42} \\
\hline
\multirow{4}{*}{2000-GOC}
  & FNN     & 74.00 & 75.00 & \textbf{27.00} & \textbf{140.00}  & 39.38 & 39.39 & 151.58 & 59.54 & 0.57 \\
  & CNN     & 121.00  & 121.00 & 57.00 & 254.00  & 37.70 & 37.70 & 114.24 & \textbf{27.90} & \textbf{0.02} \\
  & GCN     & N/A  & N/A  & N/A & N/A   & N/A & N/A & N/A & N/A & N/A \\
  & HH-MPNN & \textbf{68.28}  & \textbf{68.81}  & 39.65 & 198.00 & \textbf{30.20} & \textbf{30.14} & \textbf{100.73} & 41.86 & 0.39 \\
\hline\hline
\end{tabular}
\caption{Constraint violations and optimality gap for GridFM-datakit Dataset. The best (lowest) violations and optimality gaps are highlighted.}
\label{tab:con_vio_and_optimality_datakit}
\end{table*}
\vspace{-2mm}
\subsection{Zero-shot generalization to N-1 cases}
A key property of graph-based models is adaptability to topology changes. We demonstrate zero-shot generalization whereby a model trained only on full topology datasets (\textit{mfull}) is applied to the test subset of the N-1 dataset without retraining. We compare this zero-shot experiment with models trained on the N-1 training subset (\textit{mnminusone}) to establish an upper bound, which we refer to as the base experiment. 
\begin{figure}[t!]
    \centering
    \includegraphics[width=0.48\textwidth]{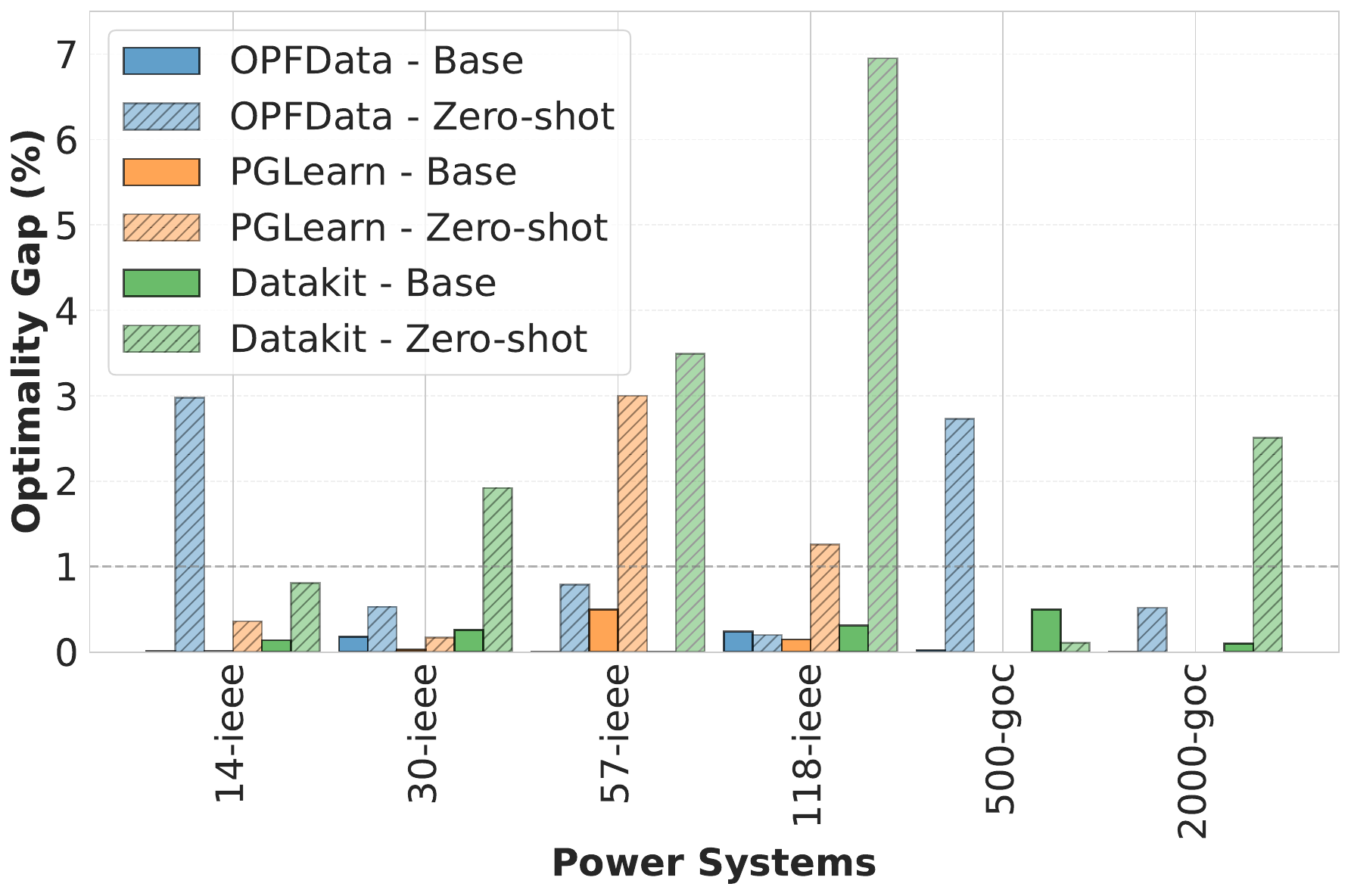}
    \vspace{-1mm}
    \caption{Comparison of average optimality gaps of the proposed HH-MPNN model in the base and zero-shot experiments. Zero-shot optimality gaps show moderate degradation with increasing system size. Zero-shot generalization is weakest for the Datakit 118-bus system.}
    \label{fig:opt_gap_compare}
    \vspace{-0.1cm}
\end{figure}

\textbf{Optimality gap (Figure \ref{fig:opt_gap_compare})}: HH-MPNN demonstrates robust zero-shot performance, maintaining  $<3\%$ optimality gap in experiments on both the OPFData and PGLearn datasets despite training only on the full topology data. This shows the model's flexibility to handle topological variations. However, results on the Datakit dataset show the fundamental limit of zero-shot generalization. Specifically, the 118-bus shows a $6.9\%$ optimality gap in the zero-shot setting. This underscores that a topology-aware architecture may still struggle when the underlying power flow patterns in unseen scenarios deviate significantly from the training distribution. We illustrate this in Fig. \ref{fig:distribution_compare}, where the distribution shift on the 118-bus datakit dataset is the most pronounced. Large voltage angles may also indicate that lines may be operating near their thermal limits, and many constraints are likely active — in other words, the out-of-distribution cases in the N-1 dataset are operationally challenging high-impact contingencies. Despite this challenge, the zero-shot optimality gaps remain comparable to those of DCOPF approximations ($3-4\%$ optimality gap)  \cite{piloto_canos_2024}, demonstrating that the model retains practical utility even under distribution shift. Interestingly, the zero-shot performance does not degrade monotonically with increasing system size; zero-shot performance in the 500-bus case, for example, is better than the 118-bus case. This suggests that the difficulty of the OPF problem may depend more on the number of active constraints in a scenario than on the grid size. As one might expect, zero-shot experiments show larger violations than the base models, due to the distribution shift between the full-topology training data and the N-1 topology test data. Constraint satisfaction can be further improved through post-processing: we demonstrate the application of power flow post-processing on the 118-bus system in Appendix \ref{appendix-B}, and note that other feasibility projection techniques may also be employed to ensure full constraint satisfaction.
\begin{figure}[t!]
    \centering
    \includegraphics[width=0.48\textwidth]{figures/distribution_comparison.png}
    \vspace{-1mm}
    \caption{Distribution of true voltage angles in the full-topology training data subset and N-1 topology testing data subset, for the 14 and 118-bus systems, across the 3 datasets.}
    \label{fig:distribution_compare}
\end{figure}
\vspace{-2mm}
\subsection{Robustness through high-impact contingency training}
While the HH-MPNN achieves stable zero-shot performance on most grids, the sharp degradation in optimality gap for some test cases suggests that zero-shot transfer has distributional limits. To address this, we develop a training strategy which uses the full topology dataset augmented by a few high-impact contingency samples from the N-1 training dataset. We selected the top 100 most expensive samples from the N-1 training dataset as measured by objective cost, representing \textless$0.1\%$ of the N-1 training subset. The model trained on this augmented full topology dataset is evaluated on the N-1 test dataset. This training approach reduced the average zero-shot optimality gap from \textbf{$6.9\%$} to \textbf{$1.4\%$} on the 118-bus Datakit dataset. Our findings reveal that exhaustive N-1 data generation is unnecessary for a topologically flexible model such as HH-MPNN; robust N-1 generalization requires only a small, targeted subset of high-impact N-1 scenarios to bridge the gap between topological flexibility and distribution shifts. Where no N-1 dataset exists, identifying high-impact contingencies for targeted training can be done using existing security analysis tools or fast linear proxies.
\vspace{-4mm}
\subsection{Size generalization}
Since HH-MPNN operates on power grids of arbitrary size, we investigate size generalization: whether models can generalize to larger unseen grids. We train a model on combined full topology Datakit training datasets from smaller (14, 30, 57, 118, and 500-bus) grids. We then test on the 2,000-bus grid in 3 experiments: (1) Training from scratch with only $5\%$ training data from the 2,000-bus grid (HH-MPNN-S). (2) Pretraining on smaller grids and testing the 2000-bus grid with no finetuning. (HH-MPNN-NF). (3) Fine-tuning the pretrained model on the same $5\%$ 2,000-bus training samples (HH-MPNN-WF).
Tables \ref{tab:model_size_gen} show that HH-MPNN-WF achieves lower average constraint violations and $64\%$ lower optimality gap (1.4\% versus 0.5\%) than HH-MPNN-S. HH-MPNN-NF leads to orders of magnitude worse predictions, showing the necessity of fine-tuning with a few samples from the target grid to learn scale-dependent features. We note that the model trained on the full training dataset still slightly outperforms HH-MPNN-WF, but HH-MPNN-WF achieves a comparable result with $95\%$ less training data. 
\begin{table}[t]
\centering
\setlength{\tabcolsep}{4pt}
\renewcommand{\arraystretch}{1.1}
\begin{tabular}{|c|c|c|c|c|c|}
\hline
\multirow{2}{*}{Model} 
& \multicolumn{4}{c|}{Violations ($\times 10^{-4}pu$)} 
& \multirow{2}{*}{Opt-gap (\%)} \\
\cline{2-5}
 & $S_{ij}(+)$ & $S_{ij}(-)$ & $P_b$ & $Q_b$ & \\
\hline
HH-MPNN-S   & 698.79  & 704.08  & 349.85 & 1595.00 & 1.40 \\
HH-MPNN-NF  & 4117.27 & 4149.95 & 3396.74 & 10953.09 & 57.44 \\ 
HH-MPNN-WF  & \textbf{275.35}  & \textbf{276.70}  &\textbf{124.21} & \textbf{575.10} & \textbf{0.47} \\
\hline
\end{tabular}
\caption{Average constraint violations and optimality gap for the proposed model trained from scratch (HH-MPNN-S), pretrained model without finetuning (HH-MPNN-NF) and pretrained model with finetuning (HH-MPNN-WF)}
\label{tab:model_size_gen}
\end{table}
\vspace{-2mm}
\subsection{Ablation studies}
\subsubsection{Impact of transformer and positional encoding}
We further investigate the effectiveness of the transformer with the effective resistance positional encoding component of the proposed architecture. To this end, we compare 3 experiments on the 118-bus full topology Datakit dataset. The first experiment uses the proposed model (HH-MPNN) as previously described. The second experiment uses the same heterogeneous message passing neural network (H-MPNN) but without the transformer and positional encoding. In the third experiment, we use the homogeneous graph representation with the topology-aware graph convolution neural network model (TAGConv) \cite{du_topology_2018}. TAGConv is a more expressive GNN model than the GCN baseline. We also increase the number of graph convolution layers to 14, to match the diameter of the 118-bus graph. In theory, this makes it possible for the model to propagate information from any node throughout the graph. We report the performance of the 3 models in Table \ref{tab:model_ablation_table}. The proposed model outperforms the other 2 models by a large margin, which shows the effectiveness of our approach. It is noteworthy that increasing the number of layers of an expressive homogeneous GNN model still gives inferior performance. This can be attributed to the well-known oversquashing problem in GNN. In contrast, the heterogeneous graph representation mitigates some of the local representation bottleneck problems, while the HH-MPNN model with transformer and effective resistance positional encoding obtains the best performance by enabling long-range information propagation throughout the graph. 
\begin{table}[b]
\centering
\setlength{\tabcolsep}{4pt}
\renewcommand{\arraystretch}{1.1}
\begin{tabular}{|c|c|c|c|c|c|}
\hline
\multirow{2}{*}{Model} 
& \multicolumn{4}{c|}{Violations ($\times 10^{-4}pu$)} 
& \multirow{2}{*}{Opt-gap (\%)} \\
\cline{2-5}
 & $S_{ij}(+)$ & $S_{ij}(-)$ & $P_b$ & $Q_b$ & \\
\hline
HH-MPNN    & \textbf{38.24}  & \textbf{39.23}  & \textbf{3.91} & \textbf{58.20} & \textbf{0.10} \\
H-MPNN   & 790.70 & 790.70 & 51.80 & 312.80 & 1.24 \\ 
TAGConv-14  & 1207.00  & 1209.00  & 23.84 & 677.50 & 5.15 \\
\hline
\end{tabular}
\caption{Average constraint violations and optimality gap for the proposed model (HH-MPNN), the same heterogeneous MPNN model without the transformer and effective resistance positional encoding (H-MPNN), and a homogeneous TAGConv model with 14 layers.}
\label{tab:model_ablation_table}
\end{table}

\subsubsection{Impact of training data size}
\begin{figure}
    \centering
    \includegraphics[width=0.48\textwidth]{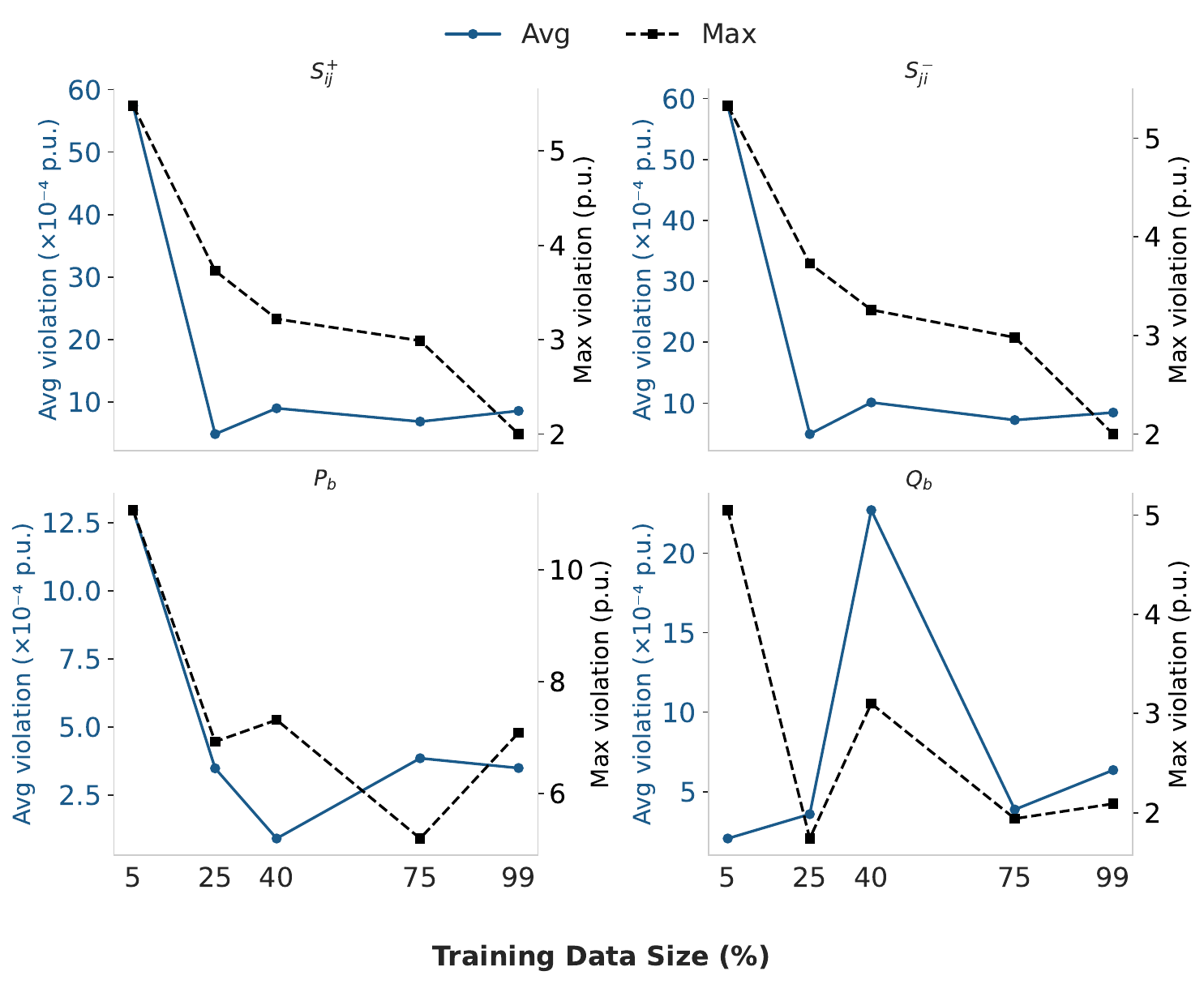}
    \caption{Average and maximum constraint violations with different training dataset sizes on the PGLearn 118-bus dataset. About 25$\%$ of available data (200,000 samples) in this case is sufficient to give a near-optimal result.}
    \label{fig:datasize_scaling}
    \vspace{-0.1cm}
\end{figure}

We also explore the impact of the training dataset size on the model performance. For this, we use the PGLearn 118-bus dataset due to the large number of available samples. The dataset is split into a training set and a testing set with approximately 800,000 and 200,000 samples, respectively. We further split the training set into a training and a validation set. We carry out 5 experiments using 5, 25, 40, 75 and 99$\%$ of the original training samples as training data and the rest as the validation set. We evaluate the trained model on the separate test dataset in each case. Fig. \ref{fig:datasize_scaling} shows the scaling behaviour of the proposed HH-MPNN model. Increasing the dataset improves constraint violations, in particular, the maximum or worst-case constraint violations. However, a moderate dataset of about $25 - 40\%$ is already sufficient for near-optimal performance with all the constraints except $Qb$, which shows sensitivity to dataset distribution. Beyond a training sample size of about 300,000, diminishing performance returns are observed. 
\vspace{-3mm}
\subsection{Analysing computational time}
Table \ref{tab:runtime_speedup} compares solution times between IPOPT and HH-MPNN. IPOPT computational time scales as $\mathcal{O}({n^{0.3-1.9}})$ with grid size, making real-time solution impractical for large grids. HH-MPNN maintains nearly constant inference time (about 13ms) across all grid sizes, providing speedups of $10^3$ to $10^4$ compared to IPOPT. Beyond inference speedups, the graph-based approach reduces dataset generation costs. Zero-shot N-1 generalization eliminates the need to generate exhaustive N-1 training data, while size generalization enables pre-training on smaller grids with lower-cost data generation, reducing dataset requirements for new large systems. For example, the 15,000-sample fine-tuning dataset requires only 350 CPU hours to generate, compared to 6,300 CPU hours for the full 270,000-sample training set, demonstrating substantial computational savings. 
\begin{table}[t]
\centering
\begin{tabular}{|c|c|c|c|}
\hline
\multicolumn{4}{|c|}{Average run time per instance $(\times 10^{-3}s$)} \\
\hline
System     & IPOPT     & HH-MPNN   & Average speed-up \\
\hline
14-ieee    & 33.2$^*$      & 12.4   & $\times$3        \\
30-ieee    & 102.3$^*$     & 12.5   & $\times$8        \\
57-ieee    & 328.8$^*$     & 12.2   & $\times$27       \\
118-ieee   & 420.7$^*$     & 12.8   & $\times$32       \\
500-goc    & 6486.0$^\dagger$    & 13.7   & $\times$473      \\
2000-goc   & 83781.0$^\dagger$   & 14.7   & $\times$5700     \\
\hline
\end{tabular}
\caption{Average run time per instance comparing IPOPT and HH-MPNN, with computed average speed-up factors.}
\label{tab:runtime_speedup}
\end{table}
\vspace{-2mm}
\section{Discussions}
The proposed HH-MPNN provides accurate but approximate ACOPF solutions across various grid sizes. For grids with fixed topology, our approach achieves performance comparable to FNN and CNN models while demonstrating the flexibility to handle N-1 topology changes that topology-specific architectures like FNN and CNN cannot. Our results challenge the assumption that graph-based models perform poorly when predicting generator setpoints and voltage angles. With sufficiently expressive architecture—combining heterogeneous graphs and transformers—both local and globally-dependent variables can be approximated accurately. We demonstrated topological flexibility through zero-shot prediction on N-1 topology variations and fine-tuning pretrained models on larger grids. We explore N-1 generalization limits, showing how distribution shifts affect model performance for operationally challenging contingency cases. Through this insight, we develop an approach for efficient and robust N-1 generalization through the inclusion of a few challenging contingency cases in the training dataset.
\footnotetext[1]{IPOPT timing results from \cite{fioretto_predicting_2020}}
\footnotetext[2]{IPOPT timing results from \cite{piloto_canos_2024}}

\textbf{Limitations}: We cannot guarantee satisfaction of all constraints. However, other feasibility restoring frameworks, such as physics-informed loss or differentiable constrained optimization \cite{nguyen_fsnet_2025, donti_dc3_2021}, may be combined with our proposed approach to satisfy all constraints. Pre-computation of effective resistance matrices can be computationally intensive for large grids. Finally, future work should explore hybrid ML-optimization approaches for feasibility guarantees and rigorous theoretical analysis of generalization bounds.
\vspace{-2mm}
\section{Conclusion}
We proposed a Hybrid Heterogeneous Message Passing Neural Network (HH-MPNN) for AC Optimal Power Flow that addresses topology adaptability and scalability challenges in existing machine learning approaches. Our architecture achieves $<1\%$ optimality gaps on challenging test cases with variable loads, generator cost, line parameters and system topologies, while maintaining constraint violations comparable to state-of-the-art models.
We show the strong generalization ability of the model through zero-shot transfer to N-1 contingency cases. Pre-training on smaller grids also improves performance on larger systems. Computational speedups of $10^{2}$ to $10^{3}$ compared to interior point solvers make the approach practical for real-time operations. These results demonstrate the potential for unified, generalizable graph-based models for diverse power system applications.
\vspace{-2mm}
\section{Acknowledgement}
This research was funded by the Dutch Research Council, Veni Talent Program Grant 19161. We thank Ali Rajaei, Balthazar Donon, Jochen Stiasny, and Megha Khosla for insightful discussions during this work. We also thank Alban Puech for help with using the GridFM-Datakit tool.
\bibliographystyle{IEEEtran}
\bibliography{references-2}

@article{cain_history_2012,
	title = {History of {Optimal} {Power} {Flow} and {Formulations}},
    journal = {Federal Energy Regulatory Commission (FERC)},
	language = {en},
	author = {Cain, Mary B and O’Neill, Richard P and Castillo, Anya},
	year = {2012},
}

@article{khaloie_review_2024,
	title = {Review of {Machine} {Learning} {Techniques} for {Optimal} {Power} {Flow}},
	issn = {1556-5068},
	doi = {10.2139/ssrn.4681955},
	language = {en},
	urldate = {2025-02-07},
	journal = {SSRN Electronic Journal},
	author = {Khaloie, Hooman and Dolanyi, Mihaly and Toubeau, Jean-Francois and Vallée, François},
	year = {2024},
}

@article{panciatici_operating_2012,
	title = {Operating in the {Fog}: {Security} {Management} {Under} {Uncertainty}},
	volume = {10},
	copyright = {https://ieeexplore.ieee.org/Xplorehelp/downloads/license-information/IEEE.html},
	issn = {1540-7977},
	shorttitle = {Operating in the {Fog}},
	doi = {10.1109/MPE.2012.2205318},
	language = {en},
	number = {5},
	urldate = {2024-08-19},
	journal = {IEEE Power and Energy Magazine},
	author = {Panciatici, Patrick and Bareux, Gabriel and Wehenkel, Louis},
	month = sep,
	year = {2012},
	pages = {40--49},
}

@article{hamann_foundation_2024,
	title = {Foundation models for the electric power grid},
	volume = {8},
	issn = {25424351},
	doi = {10.1016/j.joule.2024.11.002},
	language = {en},
	number = {12},
	urldate = {2025-02-07},
	journal = {Joule},
	author = {Hamann, Hendrik F. and et al.},
	month = dec,
	year = {2024},
	pages = {3245--3258},
}

@article{shin_accelerating_2024,
	title = {Accelerating optimal power flow with {GPUs}: {SIMD} abstraction of nonlinear programs and condensed-space interior-point methods},
	volume = {236},
	issn = {03787796},
	shorttitle = {Accelerating optimal power flow with {GPUs}},
	doi = {10.1016/j.epsr.2024.110651},
	language = {en},
	urldate = {2025-09-29},
	journal = {Electric Power Systems Research},
	author = {Shin, Sungho and Anitescu, Mihai and Pacaud, François},
	month = nov,
	year = {2024},
	pages = {110651},
}

@article{pan_deepopf_2023,
	title = {{DeepOPF}: {A} {Feasibility}-{Optimized} {Deep} {Neural} {Network} {Approach} for {AC} {Optimal} {Power} {Flow} {Problems}},
	volume = {17},
	copyright = {https://ieeexplore.ieee.org/Xplorehelp/downloads/license-information/IEEE.html},
	issn = {1932-8184, 1937-9234, 2373-7816},
	shorttitle = {{DeepOPF}},
	doi = {10.1109/JSYST.2022.3201041},
	language = {en},
	number = {1},
	urldate = {2025-02-07},
	journal = {IEEE Systems Journal},
	author = {Pan, Xiang and Chen, Minghua and Zhao, Tianyu and Low, Steven H.},
	month = mar,
	year = {2023},
	pages = {673--683},
}

@article{yang_linearized_2018,
	title = {A {Linearized} {OPF} {Model} {With} {Reactive} {Power} and {Voltage} {Magnitude}: {A} {Pathway} to {Improve} the {MW}-{Only} {DC} {OPF}},
	volume = {33},
	copyright = {https://ieeexplore.ieee.org/Xplorehelp/downloads/license-information/IEEE.html},
	issn = {0885-8950, 1558-0679},
	shorttitle = {A {Linearized} {OPF} {Model} {With} {Reactive} {Power} and {Voltage} {Magnitude}},
	doi = {10.1109/tpwrs.2017.2718551},
	language = {en},
	number = {2},
	urldate = {2025-07-14},
	journal = {IEEE Transactions on Power Systems},
	publisher = {Institute of Electrical and Electronics Engineers (IEEE)},
	author = {Yang, Zhifang and Zhong, Haiwang and Bose, Anjan and Zheng, Tongxin and Xia, Qing and Kang, Chongqing},
	month = mar,
	year = {2018},
	pages = {1734--1745},
}

@article{molzahn_implementation_2013,
	title = {Implementation of a {Large}-{Scale} {Optimal} {Power} {Flow} {Solver} {Based} on {Semidefinite} {Programming}},
	volume = {28},
	copyright = {https://ieeexplore.ieee.org/Xplorehelp/downloads/license-information/IEEE.html},
	issn = {0885-8950, 1558-0679},
	doi = {10.1109/tpwrs.2013.2258044},
	language = {en},
	number = {4},
	urldate = {2025-07-14},
	journal = {IEEE Transactions on Power Systems},
	publisher = {Institute of Electrical and Electronics Engineers (IEEE)},
	author = {Molzahn, Daniel K. and Holzer, Jesse T. and Lesieutre, Bernard C. and DeMarco, Christopher L.},
	month = nov,
	year = {2013},
	pages = {3987--3998},
}

@misc{chatzos_high-fidelity_2020,
	title = {High-{Fidelity} {Machine} {Learning} {Approximations} of {Large}-{Scale} {Optimal} {Power} {Flow}},
	doi = {10.48550/arXiv.2006.16356},
	language = {en},
	urldate = {2025-02-07},
	publisher = {arXiv},
	author = {Chatzos, Minas and Fioretto, Ferdinando and Mak, Terrence W. K. and Hentenryck, Pascal Van},
	month = jun,
	year = {2020},
	note = {arXiv:2006.16356 [eess]},
}

@inproceedings{baker_solutions_2021,
	address = {Virtual Event Italy},
	title = {Solutions of {DC} {OPF} are {Never} {AC} {Feasible}},
	isbn = {978-1-4503-8333-2},
	doi = {10.1145/3447555.3464875},
	language = {en},
	urldate = {2025-02-10},
	booktitle = {Proceedings of the {Twelfth} {ACM} {International} {Conference} on {Future} {Energy} {Systems}},
	publisher = {ACM},
	author = {Baker, Kyri},
	month = jun,
	year = {2021},
	pages = {264--268},
}

@inproceedings{zamzam_learning_2020,
	address = {Tempe, AZ, USA},
	title = {Learning {Optimal} {Solutions} for {Extremely} {Fast} {AC} {Optimal} {Power} {Flow}},
	copyright = {https://ieeexplore.ieee.org/Xplorehelp/downloads/license-information/IEEE.html},
	isbn = {978-1-72816-127-3},
	doi = {10.1109/SmartGridComm47815.2020.9303008},
	language = {en},
	urldate = {2025-02-10},
	booktitle = {2020 {IEEE} {International} {Conference} on {Communications}, {Control}, and {Computing} {Technologies} for {Smart} {Grids} ({SmartGridComm})},
	publisher = {IEEE},
	author = {Zamzam, Ahmed S. and Baker, Kyri},
	month = nov,
	year = {2020},
	pages = {1--6},
}

@article{park_compact_2024,
	title = {Compact {Optimization} {Learning} for {AC} {Optimal} {Power} {Flow}},
	volume = {39},
	copyright = {https://ieeexplore.ieee.org/Xplorehelp/downloads/license-information/IEEE.html},
	issn = {0885-8950, 1558-0679},
	doi = {10.1109/TPWRS.2023.3313438},
	language = {en},
	number = {2},
	urldate = {2025-02-07},
	journal = {IEEE Transactions on Power Systems},
	author = {Park, Seonho and Chen, Wenbo and Mak, Terrence W.K. and Van Hentenryck, Pascal},
	month = mar,
	year = {2024},
	pages = {4350--4359},
}

@article{huang_unsupervised_2024,
	title = {Unsupervised {Learning} for {Solving} {AC} {Optimal} {Power} {Flows}: {Design}, {Analysis}, and {Experiment}},
	volume = {39},
	copyright = {https://ieeexplore.ieee.org/Xplorehelp/downloads/license-information/IEEE.html},
	issn = {0885-8950, 1558-0679},
	shorttitle = {Unsupervised {Learning} for {Solving} {AC} {Optimal} {Power} {Flows}},
	doi = {10.1109/TPWRS.2024.3373399},
	language = {en},
	number = {6},
	urldate = {2025-02-07},
	journal = {IEEE Transactions on Power Systems},
	author = {Huang, Wanjun and Chen, Minghua and Low, Steven H.},
	month = nov,
	year = {2024},
	pages = {7102--7114},
}

@inproceedings{owerko_optimal_2020,
	address = {Barcelona, Spain},
	title = {Optimal {Power} {Flow} {Using} {Graph} {Neural} {Networks}},
	copyright = {https://ieeexplore.ieee.org/Xplorehelp/downloads/license-information/IEEE.html},
	isbn = {978-1-5090-6631-5},
	doi = {10.1109/ICASSP40776.2020.9053140},
	language = {en},
	urldate = {2025-04-14},
	booktitle = {{ICASSP} 2020 - 2020 {IEEE} {International} {Conference} on {Acoustics}, {Speech} and {Signal} {Processing} ({ICASSP})},
	publisher = {IEEE},
	author = {Owerko, Damian and Gama, Fernando and Ribeiro, Alejandro},
	month = may,
	year = {2020},
	pages = {5930--5934},
}

@inproceedings{baker_learning_2019,
	address = {Pittsburgh, PA, USA},
	title = {Learning {Warm}-{Start} {Points} {For} {Ac} {Optimal} {Power} {Flow}},
	copyright = {https://ieeexplore.ieee.org/Xplorehelp/downloads/license-information/IEEE.html},
	isbn = {978-1-7281-0824-7},
	doi = {10.1109/MLSP.2019.8918690},
	language = {en},
	urldate = {2026-04-20},
	booktitle = {2019 {IEEE} 29th {International} {Workshop} on {Machine} {Learning} for {Signal} {Processing} ({MLSP})},
	publisher = {IEEE},
	author = {Baker, Kyri},
	month = oct,
	year = {2019},
	pages = {1--6},
}

@inproceedings{crozier_data-driven_2022,
	address = {Denver, CO, USA},
	title = {Data-driven {Probabilistic} {Constraint} {Elimination} for {Accelerated} {Optimal} {Power} {Flow}},
	copyright = {https://doi.org/10.15223/policy-029},
	isbn = {978-1-66540-823-3},
	doi = {10.1109/PESGM48719.2022.9916838},
	language = {en},
	urldate = {2025-03-21},
	booktitle = {2022 {IEEE} {Power} \& {Energy} {Society} {General} {Meeting} ({PESGM})},
	publisher = {IEEE},
	author = {Crozier, Constance and Baker, Kyri},
	month = jul,
	year = {2022},
	pages = {1--5},
}

@article{chatzos_spatial_2022,
	title = {Spatial {Network} {Decomposition} for {Fast} and {Scalable} {AC}-{OPF} {Learning}},
	volume = {37},
	copyright = {https://ieeexplore.ieee.org/Xplorehelp/downloads/license-information/IEEE.html},
	issn = {0885-8950, 1558-0679},
	doi = {10.1109/TPWRS.2021.3124726},
	language = {en},
	number = {4},
	urldate = {2025-02-07},
	journal = {IEEE Transactions on Power Systems},
	author = {Chatzos, Minas and Mak, Terrence W. K. and Hentenryck, Pascal Van},
	month = jul,
	year = {2022},
	pages = {2601--2612},
}

@article{falconer_leveraging_2023,
	title = {Leveraging {Power} {Grid} {Topology} in {Machine} {Learning} {Assisted} {Optimal} {Power} {Flow}},
	volume = {38},
	copyright = {https://ieeexplore.ieee.org/Xplorehelp/downloads/license-information/IEEE.html},
	issn = {0885-8950, 1558-0679},
	doi = {10.1109/TPWRS.2022.3187218},
	language = {en},
	number = {3},
	urldate = {2025-02-07},
	journal = {IEEE Transactions on Power Systems},
	author = {Falconer, Thomas and Mones, Letif},
	month = may,
	year = {2023},
	pages = {2234--2246},
}

@article{jia_convopf-dop_2023,
	title = {{ConvOPF}-{DOP}: {A} {Data}-{Driven} {Method} for {Solving} {AC}-{OPF} {Based} on {CNN} {Considering} {Different} {Operation} {Patterns}},
	volume = {38},
	copyright = {https://ieeexplore.ieee.org/Xplorehelp/downloads/license-information/IEEE.html},
	issn = {0885-8950, 1558-0679},
	shorttitle = {{ConvOPF}-{DOP}},
	doi = {10.1109/TPWRS.2022.3163381},
	language = {en},
	number = {1},
	urldate = {2025-02-07},
	journal = {IEEE Transactions on Power Systems},
	author = {Jia, Yujing and Bai, Xiaoqing and Zheng, Liqin and Weng, Zonglong and Li, Yunyi},
	month = jan,
	year = {2023},
	pages = {853--860},
}

@article{zhou_deepopf-ft_2023,
	title = {{DeepOPF}-{FT}: {One} {Deep} {Neural} {Network} for {Multiple} {AC}-{OPF} {Problems} {With} {Flexible} {Topology}},
	volume = {38},
	copyright = {https://ieeexplore.ieee.org/Xplorehelp/downloads/license-information/IEEE.html},
	issn = {0885-8950, 1558-0679},
	shorttitle = {{DeepOPF}-{FT}},
	doi = {10.1109/TPWRS.2022.3217407},
	language = {en},
	number = {1},
	urldate = {2025-02-07},
	journal = {IEEE Transactions on Power Systems},
	author = {Zhou, Min and Chen, Minghua and Low, Steven H.},
	month = jan,
	year = {2023},
	pages = {964--967},
}

@inproceedings{owerko_unsupervised_2024,
	address = {Seoul, Korea, Republic of},
	title = {Unsupervised {Optimal} {Power} {Flow} {Using} {Graph} {Neural} {Networks}},
	copyright = {https://doi.org/10.15223/policy-029},
	isbn = {9798350344851},
	doi = {10.1109/ICASSP48485.2024.10446827},
	language = {en},
	urldate = {2025-03-21},
	booktitle = {{ICASSP} 2024 - 2024 {IEEE} {International} {Conference} on {Acoustics}, {Speech} and {Signal} {Processing} ({ICASSP})},
	publisher = {IEEE},
	author = {Owerko, Damian and Gama, Fernando and Ribeiro, Alejandro},
	month = apr,
	year = {2024},
	pages = {6885--6889},
}

@article{gao_physics-guided_2024,
	title = {A {Physics}-{Guided} {Graph} {Convolution} {Neural} {Network} for {Optimal} {Power} {Flow}},
	volume = {39},
	copyright = {https://ieeexplore.ieee.org/Xplorehelp/downloads/license-information/IEEE.html},
	issn = {0885-8950, 1558-0679},
	doi = {10.1109/TPWRS.2023.3238377},
	language = {en},
	number = {1},
	urldate = {2025-02-07},
	journal = {IEEE Transactions on Power Systems},
	author = {Gao, Maosheng and Yu, Juan and Yang, Zhifang and Zhao, Junbo},
	month = jan,
	year = {2024},
	pages = {380--390},
}

@misc{varbella_physics-informed_2024,
	title = {Physics-{Informed} {GNN} for non-linear constrained optimization: {PINCO} a solver for the {AC}-optimal power flow},
	shorttitle = {Physics-{Informed} {GNN} for non-linear constrained optimization},
	doi = {10.48550/arXiv.2410.04818},
	language = {en},
	urldate = {2026-02-05},
	publisher = {arXiv},
	author = {Varbella, Anna and Briens, Damien and Gjorgiev, Blazhe and D'Inverno, Giuseppe Alessio and Sansavini, Giovanni},
	month = oct,
	year = {2024},
	note = {arXiv:2410.04818 [eess]},
}

@article{lopez-garcia_optimal_2025,
	title = {Optimal {Power} {Flow} {With} {Physics}-{Informed} {Typed} {Graph} {Neural} {Networks}},
	volume = {40},
	copyright = {https://creativecommons.org/licenses/by/4.0/legalcode},
	issn = {0885-8950, 1558-0679},
	doi = {10.1109/TPWRS.2024.3394371},
	language = {en},
	number = {1},
	urldate = {2026-02-11},
	journal = {IEEE Transactions on Power Systems},
	author = {Lopez-Garcia, Tania B. and Domínguez-Navarro, José Antonio},
	month = jan,
	year = {2025},
	pages = {381--393},
}

@article{deihim_initial_2024,
	title = {Initial estimate of {AC} optimal power flow with graph neural networks},
	volume = {234},
	issn = {03787796},
	doi = {10.1016/j.epsr.2024.110782},
	language = {en},
	urldate = {2026-02-05},
	journal = {Electric Power Systems Research},
	author = {Deihim, Azad and Apostolopoulou, Dimitra and Alonso, Eduardo},
	month = sep,
	year = {2024},
	pages = {110782},
}

@article{liu_topology-aware_2023,
	title = {Topology-{Aware} {Graph} {Neural} {Networks} for {Learning} {Feasible} and {Adaptive} {AC}-{OPF} {Solutions}},
	volume = {38},
	copyright = {https://ieeexplore.ieee.org/Xplorehelp/downloads/license-information/IEEE.html},
	issn = {0885-8950, 1558-0679},
	doi = {10.1109/TPWRS.2022.3230555},
	language = {en},
	number = {6},
	urldate = {2025-02-07},
	journal = {IEEE Transactions on Power Systems},
	author = {Liu, Shaohui and Wu, Chengyang and Zhu, Hao},
	month = nov,
	year = {2023},
	pages = {5660--5670},
}

@misc{piloto_canos_2024,
	title = {{CANOS}: {A} {Fast} and {Scalable} {Neural} {AC}-{OPF} {Solver} {Robust} {To} {N}-1 {Perturbations}},
	shorttitle = {{CANOS}},
	doi = {10.48550/arXiv.2403.17660},
	language = {en},
	urldate = {2025-02-07},
	publisher = {arXiv},
	author = {Piloto, Luis and et. al},
	month = mar,
	year = {2024},
	note = {arXiv:2403.17660 [cs]},
}

@inproceedings{ghamizi_opf-hgnn_2024,
	address = {Seattle, WA, USA},
	title = {{OPF}-{HGNN}: {Generalizable} {Heterogeneous} {Graph} {Neural} {Networks} for {AC} {Optimal} {Power} {Flow}},
	copyright = {https://doi.org/10.15223/policy-029},
	isbn = {9798350381832},
	shorttitle = {{OPF}-{HGNN}},
	doi = {10.1109/PESGM51994.2024.10688560},
	language = {en},
	urldate = {2025-02-07},
	booktitle = {2024 {IEEE} {Power} \& {Energy} {Society} {General} {Meeting} ({PESGM})},
	publisher = {IEEE},
	author = {Ghamizi, Salah and Ma, Aoxiang and Cao, Jun and Rodriguez Cortes, Pedro},
	month = jul,
	year = {2024},
	pages = {1--5},
}

@misc{bau_principled_2025,
	title = {A {Principled} {Framework} to {Evaluate} {Quality} of {AC}-{OPF} {Datasets} for {Machine} {Learning}: {Benchmarking} a {Novel}, {Scalable} {Generation} {Method}},
	shorttitle = {A {Principled} {Framework} to {Evaluate} {Quality} of {AC}-{OPF} {Datasets} for {Machine} {Learning}},
	doi = {10.48550/arXiv.2508.19083},
	language = {en},
	urldate = {2026-02-11},
	publisher = {arXiv},
	author = {Baù, Matteo and Perbellini, Luca and Grillo, Samuele},
	month = aug,
	year = {2025},
	note = {arXiv:2508.19083 [eess]},
}

@misc{li_lumina_2026,
	title = {{LUMINA}: {Foundation} {Models} for {Topology} {Transferable} {ACOPF}},
	shorttitle = {{LUMINA}},
	doi = {10.48550/arXiv.2603.04300},
	language = {en},
	urldate = {2026-03-27},
	publisher = {arXiv},
	author = {Li, Yijiang and Memon, Zeeshan and Jin, Hongwei and Fenu, Stefano and Song, Keunju and Sharma, Sunash B. and Gasana, Parfait and Kim, Hongseok and Zhao, Liang and Kim, Kibaek},
	month = mar,
	year = {2026},
	note = {arXiv:2603.04300 [cs]},
}

@misc{klamkin_pglearn_2025,
	title = {{PGLearn} -- {An} {Open}-{Source} {Learning} {Toolkit} for {Optimal} {Power} {Flow}},
	doi = {10.48550/arXiv.2505.22825},
	language = {en},
	urldate = {2026-02-11},
	publisher = {arXiv},
	author = {Klamkin, Michael and Tanneau, Mathieu and Hentenryck, Pascal Van},
	month = may,
	year = {2025},
	note = {arXiv:2505.22825 [cs]},
}

@misc{puech_gridfm-datakit-v1_2025,
	title = {gridfm-datakit-v1: {A} {Python} {Library} for {Scalable} and {Realistic} {Power} {Flow} and {Optimal} {Power} {Flow} {Data} {Generation}},
	shorttitle = {gridfm-datakit-v1},
	doi = {10.48550/arXiv.2512.14658},
	language = {en},
	urldate = {2026-01-23},
	publisher = {arXiv},
	author = {Puech, Alban and et al.},
	month = dec,
	year = {2025},
	note = {arXiv:2512.14658 [cs]},
}

@misc{rampasek_recipe_2023,
	title = {Recipe for a {General}, {Powerful}, {Scalable} {Graph} {Transformer}},
	doi = {10.48550/arXiv.2205.12454},
	language = {en},
	urldate = {2026-01-23},
	publisher = {arXiv},
	author = {Rampášek, Ladislav and Galkin, Mikhail and Dwivedi, Vijay Prakash and Luu, Anh Tuan and Wolf, Guy and Beaini, Dominique},
	month = jan,
	year = {2023},
	note = {arXiv:2205.12454 [cs]},
}

@misc{battaglia_interaction_2016,
	title = {Interaction {Networks} for {Learning} about {Objects}, {Relations} and {Physics}},
	doi = {10.48550/arXiv.1612.00222},
	language = {en},
	urldate = {2025-02-10},
	publisher = {arXiv},
	author = {Battaglia, Peter W. and Pascanu, Razvan and Lai, Matthew and Rezende, Danilo and Kavukcuoglu, Koray},
	month = dec,
	year = {2016},
	note = {arXiv:1612.00222 [cs]},
}

@article{choromanski_rethinking_2021,
	address = {Austria},
	title = {Rethinking Attention with Performers},
	language = {en},
	journal = {International Conference on Learning Representations},
	author = {Choromanski, Krzysztof and et. al},
	year = {2021},
}

@article{cetinay_topological_2018,
	title = {A {Topological} {Investigation} of {Power} {Flow}},
	volume = {12},
	copyright = {https://ieeexplore.ieee.org/Xplorehelp/downloads/license-information/IEEE.html},
	issn = {1932-8184, 1937-9234, 2373-7816},
	doi = {10.1109/JSYST.2016.2573851},
	language = {en},
	number = {3},
	urldate = {2026-01-23},
	journal = {IEEE Systems Journal},
	author = {Cetinay, Hale and Kuipers, Fernando A. and Van Mieghem, Piet},
	month = sep,
	year = {2018},
	pages = {2524--2532},
}

@article{koc_impact_2014,
	title = {The impact of the topology on cascading failures in a power grid model},
	volume = {402},
	issn = {03784371},
	doi = {10.1016/j.physa.2014.01.056},
	language = {en},
	urldate = {2025-08-20},
	journal = {Physica A: Statistical Mechanics and its Applications},
	author = {Koç, Yakup and Warnier, Martijn and Mieghem, Piet Van and Kooij, Robert E. and Brazier, Frances M.T.},
	month = may,
	year = {2014},
	pages = {169--179},
}

@article{lovett_opfdata_nodate,
	title = {{OPFData}: {Large}-scale datasets for {AC} optimal power flow with topological perturbations},
	language = {en},
    month = june,
    year = {2024},
	author = {Lovett, Sean and Zgubič, Miha and Liguori, Sofia and Madjiheurem, Sephora and Tomlinson, Hamish and Elster, Sophie and Apps, Chris and Witherspoon, Sims and Piloto, Luis},
}

@misc{du_topology_2018,
	title = {Topology {Adaptive} {Graph} {Convolutional} {Networks}},
	doi = {10.48550/arXiv.1710.10370},
	language = {en},
	urldate = {2025-02-10},
	publisher = {arXiv},
	author = {Du, Jian and Zhang, Shanghang and Wu, Guanhang and Moura, Jose M. F. and Kar, Soummya},
	month = feb,
	year = {2018},
	note = {arXiv:1710.10370 [cs]},
}

@article{fioretto_predicting_2020,
	title = {Predicting {AC} {Optimal} {Power} {Flows}: {Combining} {Deep} {Learning} and {Lagrangian} {Dual} {Methods}},
	volume = {34},
	copyright = {https://www.aaai.org},
	issn = {2374-3468, 2159-5399},
	shorttitle = {Predicting {AC} {Optimal} {Power} {Flows}},
	doi = {10.1609/aaai.v34i01.5403},
	language = {en},
	number = {01},
	urldate = {2026-04-20},
	journal = {Proceedings of the AAAI Conference on Artificial Intelligence},
	author = {Fioretto, Ferdinando and Mak, Terrence W.K. and Van Hentenryck, Pascal},
	month = apr,
	year = {2020},
	pages = {630--637},
}

@article{donti_dc3_2021,
	address = {Austria},
	title = {{DC3}: {A} {Learning} {Method} {for} {Optimization} {with} {Hard} {Constraints}},
	language = {en},
	journal = {International Conference on Learning Representations},
	author = {Donti, Priya L and Rolnick, David and Kolter, J Zico},
	year = {2021},
}

@misc{nguyen_fsnet_2025,
	title = {{FSNet}: {Feasibility}-{Seeking} {Neural} {Network} for {Constrained} {Optimization} with {Guarantees}},
	shorttitle = {{FSNet}},
	doi = {10.48550/arXiv.2506.00362},
	language = {en},
	urldate = {2026-03-19},
	publisher = {arXiv},
	author = {Nguyen, Hoang T. and Donti, Priya L.},
	month = oct,
	year = {2025},
	note = {arXiv:2506.00362 [cs]},
}
\vspace{-4mm}
\appendices
\section{Problem formulation}
\label{appendix-A}
For a power system with $N$ buses, $\mathcal{E}$ branches, $G$ generators, $L$ loads and $S$ shunts, the ACOPF problem is defined as follows:
\begin{subequations}
\label{eq:AC-OPF_formulation}
\begin{align}
& \min \sum_{i \in G} a(PG_i^2) + b(PG_i) + c \label{eq:cost}\\  
& PG_i^{min} \leq PG_i \leq PG_i^{max} \quad \forall i \in G \label{eq:pgbound} \\ 
& QG_i^{min} \leq QG_i \leq QG_i^{max} \quad \forall i \in G \label{eq:qgbound} \\
& V_i^{min} \leq V_i \leq V_i^{max} \quad \forall i \in N \label{eq:vbound} \\
& \theta_{ij}^{min} \leq \theta_{ij} \leq \theta_{ij}^{max} \quad \forall (i,j) \in \mathcal{E} \label{eq:thetabound} \\
& |S_{ij}| \leq S_{ij}^{max} \quad \forall (i,j) \in \mathcal{E} \label{eq:flowbound} \\
& S_{ij} = (Y_{ij} + Y_{ij}^c)^* \cdot \frac{|V_i|^2}{|T_{ij}|^2} - Y_{ij}^* \frac{{V_i V_j^*}}{T_{ij}} \quad \forall (i,j) \in \mathcal{E} \label{eq:forwardflow} \\
& S_{ji} = (Y_{ij} + Y_{ij}^c)^* \cdot {|V_j|^2} - Y_{ij}^* \frac{V_i^* V_j}{T_{ij}^*} \quad \forall (i,j) \in \mathcal{E} \label{eq:reverseflow} \\
& \sum_{k \in G} SG_k - \sum_{k \in L} SL_k - \sum_{k \in S} SS_k = \sum_{(i,j) \in \mathcal{E}} S_{ij} \quad \forall i \in N \label{eq:powerbalance}
\end{align}
\end{subequations}
Here, $a$, $b$ and $c$ are the quadratic, linear and constant cost coefficients of generators, respectively. $PG_i$ and $QG_i$ are the generators' active and reactive powers, respectively. $V_i$ is the voltage magnitude at a bus $i$, while $\theta_{ij}$ is the voltage angle difference between buses $i$ and $j$ respectively. $S_{ij}$ is the complex power over branch $(i,j)$, in the forward direction, $S_{ji}$ in the complex branch power in reverse direction.  $SG_k$ is the complex power of generator $k$, $SL_k$ is the complex power at load $k$ and $SS_k$ is the complex power at shunt $k$.
$Y_{ij}$ is the admittance of a branch, $Y_{ij}^c$ is the charging admittance of the branch while $T_{ij}$ is the complex transformation ratio of the branch.
\vspace{-2mm}
\section{Post processing model predictions with Power Flow}
\label{appendix-B}
Power flow can be used to process the predictions from the ML models further, to ensure satisfaction of power balance constraints and make the obtained solution AC-feasible. However, we note that correcting the model predictions with power flows may now lead to violations of inequality constraints, such as voltage and power bounds. In the power flow formulation, only the slack bus has its active generation adjusted. This can lead to a violation of active power bound at the slack bus. Generations at other PV buses remain the same as predicted by the ML model. The reactive generation at PQ buses may be changed. We do not enforce reactive power limits; however, this can be achieved by converting PV buses to PQ buses at the reactive power limit. Table \ref{tab:ppf_constraint_violation} shows the changes in constraint violations and the optimality gap after power-flow postprocessing on the 118-bus grid using the Datakit dataset.
\begin{table}[h!]
\centering
\setlength{\tabcolsep}{4pt}
\renewcommand{\arraystretch}{1.2}
\resizebox{\columnwidth}{!}{%
\begin{tabular}{|l|c|c|c|c|c|c|c|c|c|}
\hline
& \multicolumn{9}{c|}{Constraint violation ($\times 10^{-4}$ p.u.)} \\
\cline{2-10}
& $S_{ij}(+)$ & $S_{ij}(-)$ & $P_b$ & $Q_b$ & $V$ & $\theta$ 
& $P_G$ & $Q_G$ & \makecell{Opt.\\gap (\%)} \\
\hline
Pre PF  & 38.24 & 39.23 & 3.91 & 58.2  & 0.00 & 0.00 & 0.00 & 0.00  & 0.10 \\
Post PF & 1.00  & 1.00  & 0.00 & 591.0 & 0.06 & 0.00 & 3.0  & 492.0 & 0.04 \\
\hline
\end{tabular}%
}
\caption{Constraint violations before and after power flow (PF).}
\label{tab:ppf_constraint_violation}
\end{table}
Post-processing with power flow reduces active power balance violations to the precision of the numerical solver. The maximum active power balance violation after post-processing is in the order of $10^{-5}pu$. However, the active and reactive generation limits may now be exceeded. The maximum active generation violation is 1.06 $pu$ while the maximum reactive generation violation is 3.97 $pu$. Power flow post-processing also significantly reduces line flow violations. In total, the power flow step only takes 466 $ms$ per scenario for the 118-bus grid, making it a scalable option for post-processing of the ML ACOPF predictions.
\vspace{-2mm}
\section{Datakit data generation}
\label{appendix-C}
Load perturbation: We use the default aggregate load profile from EIA provided by Datakit. The aggregate load profile is scaled by $ref_t$, a global scaling factor. This factor is obtained by scaling the load profile within the range $[l, u]$, where $u$ is obtained by incrementally increasing the active and reactive load from the nominal value by a factor of 0.1 until OPF does not converge. $u$ is the last factor for which opf converges. $l = (1-r)u$ where r is set to 0.4. The active and reactive load at any instance t is given by:
\begin{subequations}
\label{eq:load_perturbation}
\begin{align} 
& P_{i,t} =  P_i \times ref_t \times E_{i,t}^{p} \label{eq:p_load} \\ 
& Q_{i,t} =  Q_i \times ref_t \times E_{i,t}^{q} \label{eq:q_load} 
\end{align}
\end{subequations}
where $E_{i,t}^{p}, E_{i,t}^{q}$ are sampled from $\mathcal{U} (1-0.2, 1+0.2)$ for per-bus load variations
Admittance perturbation: Each branch reactance and resistance is sampled from $\mathcal{U} (max(0, (1-0.2)), 1+0.2)$. 
Generator cost perturbation: Each generator's cost coefficient is perturbed by scaling the base cost by a factor sampled from $\mathcal{U} (1-0.5, 1+0.5)$
N-1 topology perturbation: For the N-1 dataset, the admittance perturbation is replaced by a topology perturbation, where a generator or branch may be selected at random to be dropped.

\end{document}